\documentclass[lettersize,journal]{IEEEtran}
\usepackage{amsmath,amsfonts}
\usepackage{algorithmic}
\usepackage{array}
\usepackage[caption=false,font=normalsize,labelfont=sf,textfont=sf]{subfig}
\usepackage{textcomp}
\usepackage{stfloats}
\usepackage{url}
\usepackage{verbatim}
\usepackage{multicol}
\usepackage{multirow}
\usepackage[ruled,linesnumbered]{algorithm2e}
\usepackage{graphicx}
\def\BibTeX{{\rm B\kern-.05em{\sc i\kern-.025em b}\kern-.08em
    T\kern-.1667em\lower.7ex\hbox{E}\kern-.125emX}}
\usepackage{balance}
\usepackage{pifont}
\usepackage{booktabs}
\usepackage[breaklinks,colorlinks]{hyperref}
\begin{document}
\title{Adversarial Training: A Survey}
\author{
Mengnan Zhao,
Lihe Zhang,
% Yuqiu Kong,
Jingwen Ye,
Huchuan Lu, \textit{Fellow, IEEE},\\
Baocai Yin,
Xinchao Wang, \textit{Senior Member, IEEE}

\thanks{Manuscript received 19, October, 2024.}
\thanks{This work was supported by the National Natural Science Foundation of China under Grant 62431004 and 62276046 and by Dalian Science and Technology Innovation Foundation under Grant 2023JJ12GX015.}
\thanks{Mengnan Zhao and Baocai Yin are with the School of Computer Science and Technology, Dalian University of Technology, Dalian 116024, China. E-mail: gaoshanxingzhi@163.com, ybc@dlut.edu.cn.}
\thanks{Lihe Zhang and Huchuan Lu are with the School of Information and Communication Engineering, Dalian University of Technology, Dalian 116024, China. E-mail: \{zhanglihe, lhchuan\}@dlut.edu.cn.}
% \thanks{Yuqiu Kong is with the School of Innovation and Entrepreneurship, Dalian University of Technology, Dalian 116024, China. E-mail: yqkong@dlut.edu.cn.}
\thanks{Jingwen Ye and Xinchao Wang are with the School of Electrical and Computer Engineering, National University of Singapore, Singapore 119077. E-mail: jingweny@nus.edu.sg, xinchao@nus.edu.sg.}
}

\markboth{Journal of \LaTeX\ Class Files,~Vol.~18, No.~9, September~2020}%
{How to Use the IEEEtran \LaTeX \ Templates}

\maketitle

\begin{abstract}
% Recent studies have exposed the vulnerability of deep neural networks to adversarial attacks. 
% In response, adversarial training (AT) was proposed and has attracted increasing research interest. 
% AT refers to incorporate adversarial examples, which are generated based on the model weights and inputs at each training iteration, into the training data. 
% This survey first outlines the formulations and practical applications of AT. 
% Next, a comprehensive review of AT techniques is provided from three aspects: data enhancement, network design, and training detail setting. 
% Finally, we discuss common challenges in AT and propose promising research directions to further advance the field.
Adversarial training (AT) refers to integrating adversarial examples — inputs altered with imperceptible perturbations that can significantly impact model predictions — into the training process.
Recent studies have demonstrated the effectiveness of AT in improving the robustness of deep neural networks against diverse adversarial attacks.
However, a comprehensive overview of these developments is still missing.
This survey addresses this gap by reviewing a broad range of recent and representative studies.
Specifically, we first describe the implementation procedures and practical applications of AT, followed by a comprehensive review of AT techniques from three perspectives: data enhancement, network design, and training configurations. 
Lastly, we discuss common challenges in AT and propose several promising directions for future research. 

\end{abstract}


\section{Introduction}
Adversarial training (AT) has attracted widespread attention for its efficacy in enhancing the network robustness against perturbations \cite{reyes2024enhancing,huang2024pointcat}, which has been successfully applied in various fields such as medical image segmentation \cite{hanif2023frequency,wang2023self}, autonomous driving \cite{shibly2023towards}, and anomaly detection \cite{zhu2022adversarial,lin2024integrating}.
Specifically, AT is typically framed as a min-max optimization problem.
The outer minimization adjusts model weights to correctly classify both clean and adversarial examples, while the inner maximization generates adversarial examples by perturbing clean inputs under the fixed model weights \cite{zhao2023fast}. 
Depending on the accessibility of fixed models, such as model parameters and gradients,
these adversarial examples can be crafted by white-box attacks  \cite{kurakin2018adversarial} or black-box attacks \cite{andriushchenko2020square}.

In addition to generating adversarial data, both types of attacks can evaluate the adversarial robustness of trained models during inference \cite{tramer2019adversarial}.
Researchers also assess models with their prediction accuracy on clean samples. 
These evaluation metrics are affected by multiple factors such as attack intensity and model architecture. 
Based on these factors, we categorize popular AT approaches from three dimensions: data enhancement, network design, and training configurations.

% Furthermore, adversarial examples and their corresponding clean examples are employed to assess the performance of trained models, .
% Adversarial examples are also employed to evaluate the adversarial robustness of trained models, along with the classification accuracy on clean samples (clean-acc).

\textit{Data Perspective:} 
This aspect focuses on increasing data diversity through various data augmentation (DE) techniques, including source DE, generic DE, and adversarial DE.

Source DE primarily employs two strategies: data collection and data generation. The former typically integrates additional unlabeled samples into the training set \cite{raghunathan2019adversarial}, while the latter incorporates data generated by methods like diffusion models \cite{yang2024structure, chen2023advdiffuser}. Unlike the pre-training augmentation of source DE, generic DE augments data during training. 
For instance, techniques such as Cutout \cite{devries2017improved} and CutMix \cite{yun2019cutmix} modify local image regions through simple replacement operations. 
Data reconstruction approaches, on the other hand, reconstruct distinct samples from input representations \cite{esser2021taming}.

Furthermore, adversarial DE utilizes various adversarial attacks to generate adversarial training data. For instance, conventional AT \cite{kinfu2022analysis} employs multi-step attacks \cite{madry2017towards}, while fast AT \cite{pan2024adversarial} uses single-step attacks \cite{goodfellow2014explaining, cheng2021fast}. Beyond standard adversarial attacks, specialized strategies for AT have been developed, such as filtering out data with low attack levels \cite{zhao2023fast} or low gradient values \cite{huang2023fast} by calculating adjacent loss differences or back-propagated gradient values. Additionally, incorporating different perturbation initialization methods -- such as random initialization \cite{wong2020fast}, learnable initialization \cite{jia2022adversarial}, and prior-guided initialization \cite{jia2022prior} -- along with adjusting attack intensity \cite{kang2021understanding, xu2022a2}, can further improve data diversity.

\textit{Model Perspective:} 
This aspect describes diverse network architectures and components used in AT.

AT is applicable to different deep learning architectures, such as Convolutional Neural Networks (CNNs) \cite{xu2023probabilistic}, Graph Neural Networks (GNNs) \cite{yang2021graph}, Recurrent Neural Networks (RNNs) \cite{liu2023robust}, Transformers \cite{zhang2023transferable}, and diffusion models \cite{karras2024analyzing},
as its framework is not limited to specific network designs \cite{wong2020fast}.
Researchers also explore multi-model AT techniques \cite{deng2024understanding}, such as ensemble AT \cite{liu2022mutual}, federated AT \cite{luo2021ensemble}, and robustness 
transfer via fine-tuning \cite{patel2023learning} or knowledge distillation  \cite{ham2024neo}.
Moreover, AT performance is highly influenced by various network components \cite{singla2021low, singh2024revisiting}, \(e.g.\), activation functions \cite{xie2020smooth}, batch normalization layers \cite{xieintriguing}, dropout configurations \cite{herrmann2022pyramid}, network depth \cite{xu2021robust}, and Softmax temperature values \cite{ hou2023improving}.
% Hence, this part also reviews studies that analyze the impact of network components. 

% The third aspect involves various training details, including loss functions, labels, and optimization algorithms.
\textit{Training Configurations Perspective:}
In addition to data enhancement and network design, specifying training configurations such as loss functions, labels, and weight settings is essential for AT, \(e.g.\), improving AT stability \cite{de2020stability}.

% Specifically, loss functions vary across different AT tasks.
Specifically, the loss function should be adjusted according to the AT task.
Popular AT tasks include conventional AT, fast AT, federated AT, virtual AT, fine-tuned AT, robust distillation, curriculum AT, and domain AT. 
Common loss functions used in these tasks include cross-entropy loss \cite{lecun1998gradient}, norm-based feature regularization \cite{tong2024taxonomy}, gradient regularization \cite{andriushchenko2020understanding}, weight regularization \cite{ding2022snn}, perceptual loss (LPIPS) \cite{zhang2018unreasonable}, Kullback-Leibler (KL) divergence \cite{huang2023boosting}.
Besides, labels in loss functions for supervising adversarial examples are often modified, using techniques like label smoothing \cite{rangwani2022closer}, label interpolation \cite{song2024regional}, and label distillation \cite{zi2021revisiting}. 

Based on the constructed objective function, various weight-related strategies, including adversarial weight perturbation \cite{wu2020adversarial}, random weight perturbation \cite{hendrycks2021many}, weight standardization \cite{pang2020boosting}, and weight selection \cite{li2022subspace}, are introduced.
To optimize model weights both effectively and efficiently, researchers further adjust the Softmax temperature values \cite{wang2024out}, aggregation algorithms \cite{tolpegin2020data}, and learning rate schedules \cite{ding2023using,smith2017cyclical}.

Additionally, this work provides a detailed analysis of four major challenges faced when applying AT techniques in real-world scenarios: catastrophic overfitting, fairness, performance trade-offs, and time efficiency. For each challenge, we propose potential research directions.

In summary, this work provides a comprehensive review of contemporary AT techniques.
The main contributions of the paper are as follows:
\begin{itemize}
    \item We abstract the key factors of AT approaches and construct a unified algorithmic framework for AT.
    \item Based on these abstracted factors, we provide a systematic taxonomy of AT methods, identifying three mainstream categories: data enhancement, network design, and training configurations.
    \item We discuss critical challenges in AT and introduce promising directions for future research.
\end{itemize}

The structure of this paper is organized as follows: Section \ref{sec2} presents the formulations and applications of AT. Section \ref{sec3}, the core of this paper, provides a comprehensive overview of existing AT methods.
Section \ref{sec4} outlines key challenges in AT and presents potential solutions.
% Finally, we conclude this survey.
% Given the vast number of AT studies, this survey focuses on representative works that are recent, typical, or uniquely significant.

% wait for uodate
% includes evaluation results for several research directions in AT to support future studies. The relevant code will be made publicly available through the provided link. 

\section{Adversarial Training: Expression and Application}\label{sec2}

\subsection{Formulation of AT}
Recent AT techniques are commonly formulated as a min-max optimization issue \cite{madry2017towards}, which minimizes the worst-case loss within a specified perturbation set for each training example, expressed as:
\begin{equation}\label{eqinint}
    \min_\theta \mathbb{E}_{(x,y) \sim \mathcal{D}} \left[ \max_{\delta \in \mathcal{B}(x, \epsilon)} \ell(x + \delta, y; \theta) \right]. 
\end{equation}
Here, \(\theta\) denotes the model parameters, and \((x, y)\) represents training data sampled from the data distribution \(\mathcal{D}\). 
\(\ell(x + \delta, y; \theta)\) refers to the loss value (\(e.g.\), cross-entropy loss) calculated with adversarial examples \(x + \delta\) and their true labels \(y\). 
\(\delta\) means adversarial perturbations that are imperceptible to humans but significantly degrade model performance. 
The allowed perturbation set \(\mathcal{B}(x, \epsilon)\) is defined as:
\[ \mathcal{B}(x, \epsilon) = \{\delta\mid x + \delta \in [0, 1],\|\delta\|_p \leq \epsilon \}, \]
where \(\epsilon\) is the maximum perturbation magnitude. \(\|\delta\|_p\) quantifies the perturbation size using the \(p\)-norm. All pixels are normalized to the range of \([0, 1]\).

\begin{figure}
    \centering
    \includegraphics[width=1\linewidth]{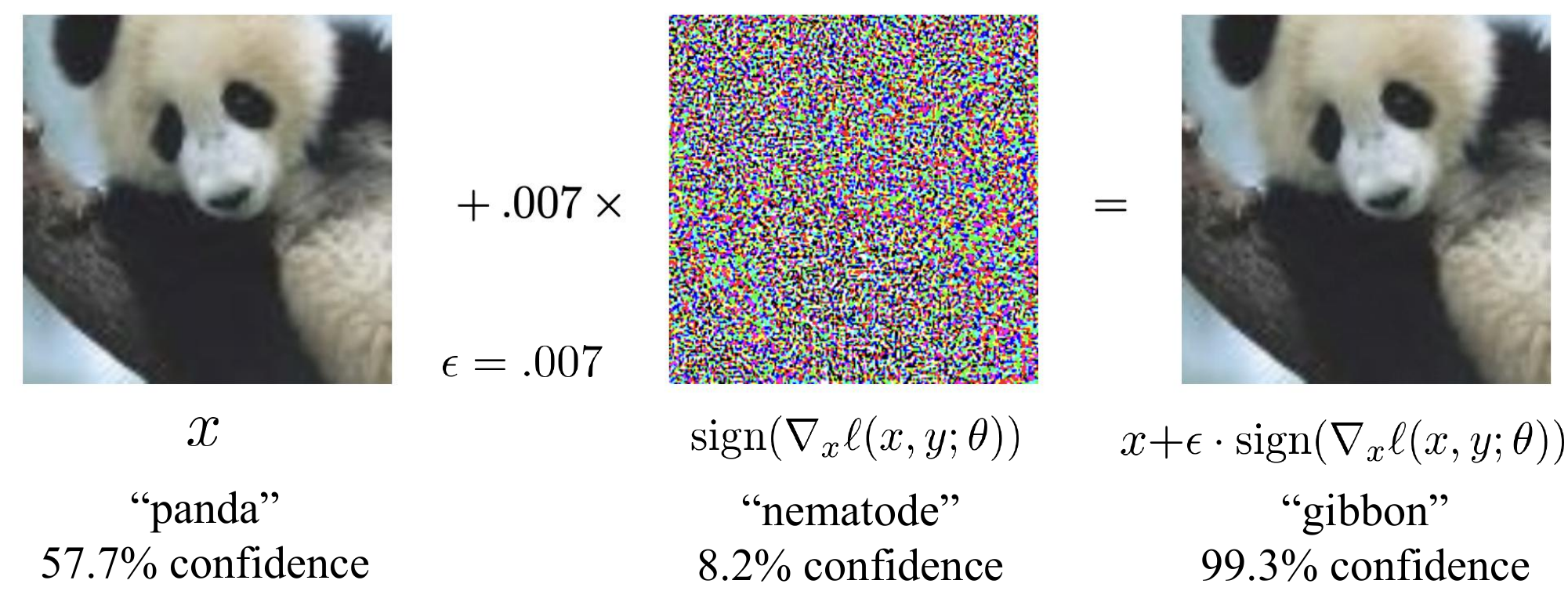}
    \caption{The adversarial perturbation \(\epsilon \cdot \text{sign}(\nabla_{x} \ell(x, y;\theta))\) causes the object `panda' to be misclassified as `gibbon'.}
    \label{fgsm}
\end{figure}

The inner maximization of Eq. (\ref{eqinint}) aims to identify \(\delta\) under fixed parameters \(\theta\). One representative method to achieve this is the Fast Gradient Sign Method (FGSM) \cite{goodfellow2014explaining}, which is a single-step adversarial attack and formulated as follows:
\[ \delta = \text{Proj}_{\mathcal{B}(x,\epsilon)} \left( \epsilon \cdot \text{sign}(\nabla_{x} \ell(x, y;\theta)) \right) \]
where \(\text{Proj}_{\mathcal{B}(x, \epsilon)}\) is the projection operator, ensuring that \(\delta\) remains within the allowed perturbation range.
The term sign(\(\cdot\)) denotes the sign function.
\(\nabla_{x}\ell\) represents the back-propagated gradient of the loss \(\ell\) with respect to the input \(x\).
Figure \ref{fgsm} provides an example of FGSM.
Furthermore, a widely-used multi-step maximization technique is Projected Gradient Descent (PGD)
% Another widely used maximization technique employs multi-step adversarial attacks such as Projected Gradient Descent (PGD) 
\cite{madry2017towards}, 
% expressed as:
\[ x_{t+1}^\prime = \text{Proj}_{\mathcal{B}(x, \epsilon)} \left( x_t^\prime + \alpha \cdot \text{sign} \left( \nabla_{x_t^\prime} \ell(x_t^\prime, y; \theta) \right) \right), \]
where \(x_t^\prime\) denotes the adversarial example at iteration \(t\). The initial adversarial example is \(x_0' = x + \delta_0\), where \(\delta_0\) represents randomly initialized perturbations. \(\alpha\) is the step size for each attack iteration. The outer minimization of Eq. (\ref{eqinint}) optimizes 
% the model parameters 
\(\theta\) on generated adversarial examples, typically using gradient-based methods such as stochastic gradient descent \cite{bottou2010large}.

\subsection{Applications of AT}
AT has been successfully applied to various tasks.

$\circ$ \textit{Abnormal event detection in videos.}
To tackle the lack of abnormal samples in training data, Georgescu et al. \cite{georgescu2021background} propose an adversarial learning strategy, which generates out-of-domain pseudo-abnormal samples using autoencoders and trains binary classifiers to distinguish between normal and abnormal latent features.

$\circ$ \textit{Biomedical image segmentation.} By conducting experiments on various eye vasculature segmentation datasets, Javanmardi et al. \cite{javanmardi2018domain} demonstrate the effectiveness of AT in mitigating the impact of inaccurately annotated segmentation targets on model performance.

$\circ$ \textit{Emitter identification.}
To enhance the discriminability of radio frequency signals while reducing sensitivity to noise, Xie et al. \cite{xie2022virtual} design a semi-supervised AT algorithm for emitter identification based on bispectrum analysis.

$\circ$ \textit{Graph embedding learning.} This transforms graph elements into representations, facilitating subsequent graph analytic tasks like node and link prediction. 
To improve graph embeddings, Pan et al. \cite{pan2019learning} enforce latent representations of adversarial data to align with a prior Gaussian distribution.

$\circ$ \textit{Healthcare debiasing.}
Biases present in healthcare data collection, stemming from variations in health conditions, can substantially affect model performance. Yang et al. \cite{yang2023adversarial} propose using AT techniques to mitigate site-specific (hospital) and demographic (ethnicity) biases in training data.

$\circ$ \textit{Human pose estimation (HPE).}
HPE localizes human anatomical keypoints and is applicable to various tasks such as action recognition and video surveillance. 
To address the vulnerability of HPE models to data corruptions like blur and pixelation, Wang et al. \cite{wang2021human} optimize the pose estimator on
more challenging samples, which are constructed by mixing different corrupted images in the generator.
% to enhance its robustness.

$\circ$ \textit{Knowledge tracing.}
By tracking the understanding degree of learners to various concepts over time, researchers aim to predict the future development of learners and identify areas requiring additional support. Guo et al. \cite{guo2021enhancing} introduce an AT-based knowledge tracing method using an attentive-LSTM backbone, which enhances generalization performance and mitigates overfitting on small datasets.

$\circ$ \textit{Large language models (LLMs).}
Pre-trained LLMs like BERT \cite{devlin2018bert} have shown impressive performance in various text-related tasks. 
For improving the adversarial robustness of these pre-trained LLMs,
Liu et al. \cite{liu2020adversarial} propose a general AT approach that regularizes the training objective by applying perturbations to the embedding space.

$\circ$ \textit{Machine reading comprehension.} Liu et al. \cite{liu2020robust} utilize a pretrained generative model to dynamically generate adversarial examples and refine model parameters for machine reading comprehension within these adversarial examples.

$\circ$ \textit{Malware detection.} 
% This task focuses on distinguishing malicious executable files from benign ones. 
Lucas et al. \cite{lucas2023adversarial} develop an efficient adversarial example generation method and examine the effectiveness of AT in mitigating various malware attacks.

$\circ$ \textit{Multi-modal sentiment analysis.} Yuan et al. \cite{yuan2023noise} propose an AT framework for multi-modal sentiment analysis, addressing issues like missing modalities and adversarial vulnerability. The framework utilizes semantic reconstruction supervision to learn unified representations for both clean and noisy data.

$\circ$ \textit{Price prediction.} Stock prices fluctuate continuously over time. To mitigate overfitting caused by static metrics like closing prices, Feng et al. \cite{feng2019enhancing} apply AT techniques to enhance the diversity of price curves.

$\circ$ \textit{Robotic learning.} Lechner et al. \cite{lechner2021adversarial} identify three key challenges in applying AT to robotic learning: (1) neural controllers may become unstable or malfunction during transitional states, (2) controllers can consistently produce errors under certain system conditions, and (3) controllers may exhibit erroneous behavior in specific environments.

$\circ$ \textit{Text to speech.}
Li et al. \cite{li2024styletts} construct a text-to-speech AT model that integrates style diffusion, AT, and large speech language models. They  employ texts from LibriTTS \cite{kawamura2024libritts} as out-of-distribution samples and transfer knowledge from speech language model encoders to generative models.

Due to page limitations, we provide an overview of 100 mainstream deep learning tasks and their corresponding AT techniques in the supplementary material.

\begin{figure*}
    \centering
    \includegraphics[width=1\linewidth]{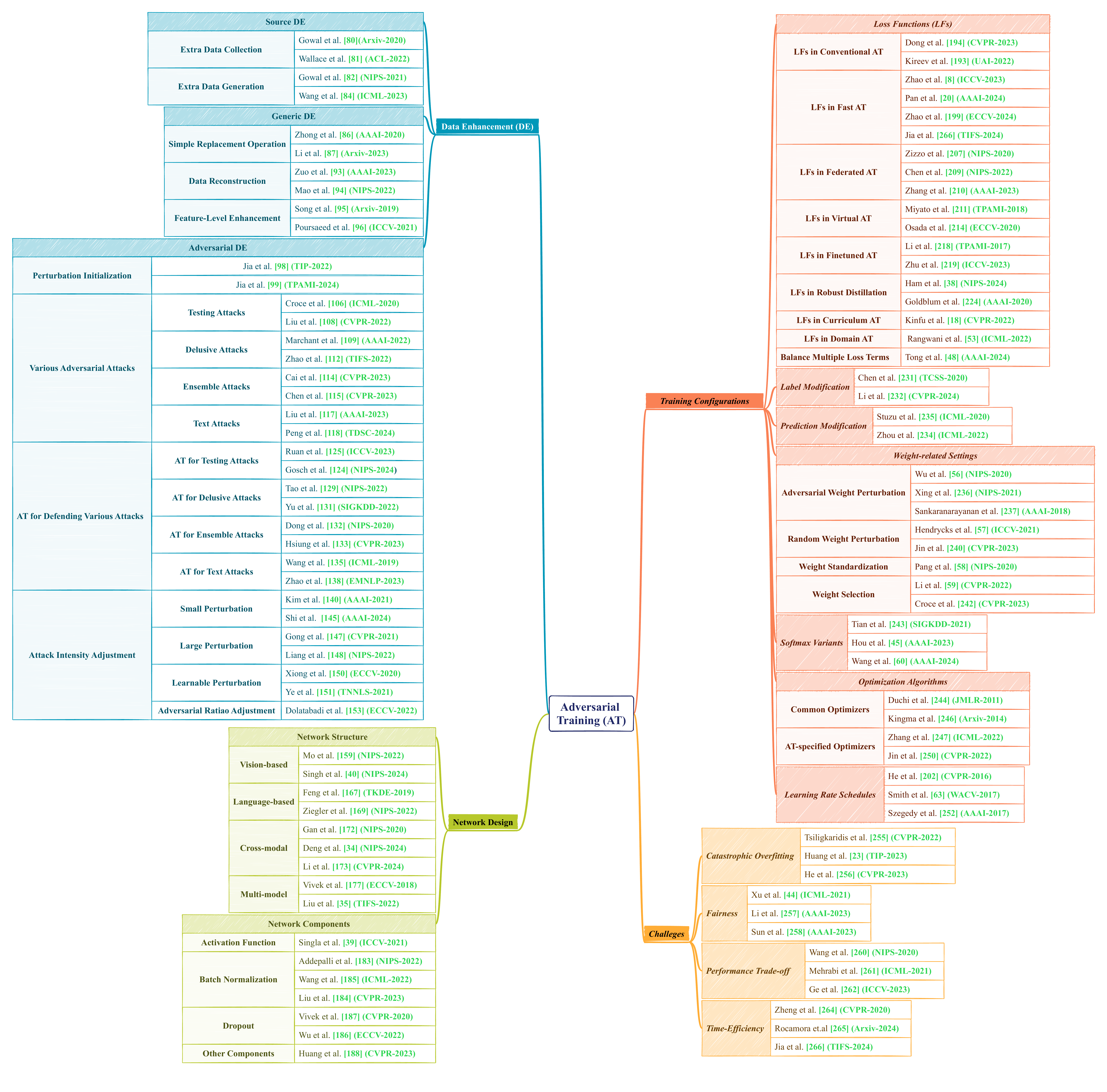}
    \vspace{-6mm}
    \caption{A comprehensive overview of adversarial training.}
    \label{AT_overall}
    \vspace{-2mm}
\end{figure*}

%为了进一步理解对抗训练过程的细节以及探索影响对抗鲁棒性能的因素，下面尽可能涵盖了对抗训练过程中可调节变量。

%数据来源\(D\)，噪声添加位置 \(P(\delta)\)，
\begin{algorithm}[t]
	\caption{Adversarial Training (AT).}
	\label{alg:algorithm1}
	\KwIn{The original dataset $\mathcal{D}$, the data enhancement technique DE($\cdot$), 
 the perturbation initialization method Init($\cdot$), 
 the label modification functions LM$_{attack}$($\cdot$) and LM$_{AT}$($\cdot$), 
 the adversarial attack approach Attack($\cdot$),
 the loss functions $\ell_{attack}(\cdot)$ and $\ell_{AT}(\cdot)$, 
 the victim model $f_{attack}(\cdot; \theta_{attack})$, 
 the Softmax function \(\sigma(\cdot)\) with the temperature \(\tau\),
 the step size of each attack iteration $\alpha$, the perturbation budget $\epsilon$, the maximum attack step $S$, 
 % the projection operator Proj($\cdot$),
 % the available perturbation positions $p_\delta$, 
 the AT model $f_{AT}(\cdot; \theta_{AT})$, 
 the optimization algorithm OP, the learning rate schedule $lr$.}
	\KwOut{The adversarially trained model $f_{AT}(\cdot; \theta_{AT})$.}  
	\BlankLine
        
	\For{$t$ \text{in} N}{
            
            $\mathcal{D}_{de}$ = DE($\mathcal{D}$);

		\For{($x_0$, $y$) $\text{in}$ $\mathcal{D}_{de}$}{

        $\delta_0$ = $\text{Init}(x_0)$;

        $y_{attack}$ = LM$_{attack}$($y$);
        
        $\delta = \text{Attack}(
\ell_{attack}(\cdot), f_{attack}(\cdot;\theta_{attack}),$ $\sigma(\cdot;\tau), x_0, \delta_0, y_{attack},\alpha,\epsilon,S)$;

        % $\delta_{proj}$ =Proj($\delta$, $p_{\delta}$, $\epsilon$);

        $x^\prime$ = $x_0$ + $\delta$;

        $y_{AT}$ = LM$_{AT}$($y$);
        
        $\ell_{op}$ = $\ell_{AT}(\sigma(f_{AT}(x^\prime;\theta_{AT}); \tau), y_{AT})$;

        $\theta_{AT}$ = OP($\ell_{op}$, $\theta_{AT}$, $lr$);

		}
        }
\end{algorithm}

\section{Overview of adversarial training}\label{sec3}
This section provides a comprehensive review of AT techniques. 
Instead of focusing on specific approaches, Algorithm \ref{alg:algorithm1} summarizes the general procedures for AT.
Furthermore, Figure \ref{AT_overall} presents a comprehensive overview of advancements in AT across three dimensions: data enhancement, network design, and training configurations.
% Table \ref{tab1} outlines the variables used in this algorithm, along with relevant research directions and corresponding methods.
% The following subsections illustrate these variables from three perspectives: data, network architecture, and training configurations.
% Notably, some AT methods may modify multiple variables. To avoid redundancy, we group these methods according to their most significant variables.

\subsection{Data Perspective}

Insufficient training data can impact AT performance, such as the trade-off between classification accuracy on clean and adversarial examples \cite{sehwag2021robust}.
Hence, researchers introduce various data enhancement (DE) techniques, primarily including source DE, generic DE and adversarial DE.

\underline{\textit{1) Source DE:}}
Source DE refers to expanding the dataset before training begins.

$\circ$ \textit{ Extra data collection.} Raghunathan et.al \cite{raghunathan2019adversarial} augment the training set with additional unlabeled data. Gowal et al. \cite{gowal2020uncovering} indicate that employing a 3:7 ratio of unlabeled to labeled data enhances adversarial robustness.
Wallace et al. \cite{wallace2021analyzing} demonstrate that multiple rounds of dynamic adversarial data collection can improve AT performance.

$\circ$ \textit{Extra data generation.} 
Collecting large datasets is time-consuming and expensive. 
Consequently, Gowal et al. \cite{gowal2021improving} utilize generative models, such as Denoising Diffusion Probabilistic Models (DDPM) \cite{ho2020denoising}, to generate extra training data. 
Additionally, Wang et al. \cite{wang2023better} show that leveraging improved diffusion models \cite{karras2022elucidating} can further augment AT performance.

\underline{\textit{2) Generic DE:}} 
After constructing the source dataset, various generic DE techniques are applied during training.

$\circ$ \textit{Simple replacement operation.} 
This category typically generates variants of the input by applying straightforward operations.
For instance, Cutout \cite{devries2017improved} sets pixel values of randomly selected image regions to zero, while Rand-Erasing \cite{zhong2020random} replaces selected pixel values with either the mean pixel value or a random value. 
CutMix \cite{yun2019cutmix} replaces deleted image regions with corresponding regions from another image and adjusts the label according to the area ratio of the two images.
CropShift \cite{li2023data} randomly crops an image and fills the remaining area with zeros.
MixUp \cite{zhang2017mixup} overlays two images at each pixel based on a pre-defined ratio and assigns labels using this ratio. 
AutoAugment \cite{cubuk2018autoaugment} introduces a search algorithm to dynamically determine the optimal data augmentation strategy.
RandAugment \cite{cubuk2020randaugment} significantly simplifies the search complexity of data augmentation policies in AutoAugment.
Additionally, texture-debiased augmentation 
\cite{hermann2020origins} applies color distortion and a less aggressive random cropping strategy to reduce the model bias towards image textures.
% CropShift \cite{li2023data}, a data augmentation method specifically designed for AT, involves randomly cropping an image and then randomly placing this cropped section into a region of the same shape as the original image, filling the remaining area with zeros.

$\circ$ \textit{Data reconstruction.} 
% In addition to employing simple combinations for data augmentation, 
Researchers also utilize advanced generative techniques to reconstruct training data. For example, Akçay et al. \cite{akcay2019ganomaly} adopt an encoder-decoder framework to project inputs into a low-dimensional latent space before reconstructing them. Zuo et al. \cite{zuo2023generative} reconstruct inputs by inpainting their masked regions. 
Mao et al. \cite{mao2022enhance} utilize Vector Generative Adversarial Network (VQGAN) \cite{esser2021taming} to generate adversarial inputs, \textit{i.e.}, calculating the worst-case perturbations for samples reconstructed from discrete representations.

$\circ$ \textit{Feature-level enhancement.}
Unlike the above input-level enhancement, feature-level enhancement directly enhances feature diversity.
For instance, Song et al. \cite{song2019robust} split inputs into multiple patches, perform row-wise and column-wise shuffling, and extract features that ensure prediction consistency before and after shuffling.
Poursaeed et al. \cite{poursaeed2021robustness} employ generative models with decoupled latent spaces, such as Style-GAN \cite{karras2019style}, to generate latent representations with different levels of variation — low, medium, and high -- on which adversarial augmentation techniques are applied.
% . Adversarial augmentation techniques are then applied to these representations. 

\underline{\textit{3) Adversarial DE:}}
Adversarial DE forms the foundation of AT techniques, focusing on designing various adversarial attacks to generate adversarial examples.
 
$\circ$ \textit{Perturbation Initialization.} Before applying attacks, incorporating initial perturbations to clean inputs can significantly improve the diversity of adversarial perturbations.
For instance, FGSM-AT utilizes uniformly random noise as initial perturbations \cite{goodfellow2014explaining}. 
Jia et al. \cite{jia2022boosting} propose a two-step perturbation initialization process, namely, refining the perturbations generated by FGSM with an initialization generator. 
Furthermore, they \cite{jia2022prior,jia2024improving} employ three types of initial perturbations to compute subsequent adversarial perturbations: perturbations from previous batches, perturbations from previous epochs, and perturbations based on momentum statistics.  
The adversarial attack in AT using the first type of initial perturbations is formulated as
\[
\delta_{b+1} = \delta_{b} + \alpha \cdot \text{sign} \left( \nabla_{\delta_{b}} \ell(x_{b+1} + \delta_{b}, y; \theta) \right),
\]
where \(\delta_b\) and \(\delta_{b+1}\) are adversarial perturbations for inputs \(x_b\) and \(x_{b+1}\), respectively. 
Similarly, for the second initialization type, the attack is expressed as
\[
\delta_{e+1} = \delta_{e} + \alpha \cdot \text{sign} \left( \nabla_{\delta_e} \ell(x + \delta_{e}, y; \theta) \right),
\]
where \(\delta_e\) and \(\delta_{e+1}\) represent adversarial perturbations for the same input at epochs \(e\) and \(e+1\), respectively.  
Additionally, the adversarial attack using momentum-based perturbation initialization is implemented as
\[
g = \text{sign}(\nabla_{\eta_{e}}\ell(x+\eta_{e},y;\theta)),
\]
\[
g_{e+1} = \mu \cdot g_{e} + g,
\]
\[
\delta_{e+1} = \text{Proj}_{\mathcal{B}(x,\epsilon)} (\eta_{e} + \alpha \cdot g),
\]
\[
\eta_{e+1} = \text{Proj}_{\mathcal{B}(x,\epsilon)}(\eta_{e} + \alpha \cdot \text{sign}(g_{e+1})),
\]
where \(g\) denotes the signed gradient, \(\eta_e\) represents the initial perturbation for epoch \(e+1\), \(g_{e+1}\) means the gradient momentum, \(\mu\) signifies the decay factor, and \( \delta_{e+1} \) is the generated adversarial perturbation.

% After constructing the initial perturbations, we explore existing methods for generating adversarial perturbations, including those from various categories such as white-box attacks, black-box attacks, ensemble attacks, and text attacks. 
$\circ$ \textit{Various adversarial attacks.} After generating the initial adversarial perturbations, we illustrate several types of adversarial attacks that further refine these perturbations.
% , such as testing and delusive attacks.

% various attacks, , are applied in AT.
% that update adversarial perturbations. This diversity arises because the inner maximization in AT does not specify the attack mode.

\textit{Testing attacks.}
Testing attacks are widely employed to assess the robustness of pre-trained models during the inference stage, including both white-box and black-box attacks.
White-box attacks, such as FGSM and PGD, assume that attackers can access the model architecture, parameters, and gradients. Similar to PGD, the basic iterative method \cite{kurakin2018adversarial} applies FGSM iteratively with a small perturbation stride to increase attack effectiveness. The momentum iterative FGSM \cite{dong2018boosting} incorporates momentum statistics to boost attack efficiency. Additionally, DeepFool \cite{moosavi2016deepfool} computes minimal perturbations required to mislead pretrained models, while the Jacobian-based saliency map attack \cite{papernot2016limitations} leverages the Jacobian matrix to identify and modify input features that significantly affect model outputs.
In contrast to these gradient-based attacks, the C\&W attack \cite{carlini2017towards} optimizes adversarial perturbations by minimizing various norm functions. The universal attack \cite{shafahi2020universal} learns an universal adversarial perturbation for multiple inputs.

Different from white-box attacks, black-box attacks rely solely on input-output queries to produce adversarial examples. For instance, Square Attack \cite{andriushchenko2020square} progressively increases the perturbation degree until the perturbed sample successfully misleads the target model. Fast Adaptive Boundary attack (FAB) \cite{croce2020minimally} introduces a search strategy that maintains adversarial effectiveness while reducing the perturbation degree.

Furthermore, AutoAttack \cite{croce2020reliable} integrates both white-box and black-box attacks, incorporating PGD variants—APGD-CE and APGD-T \cite{croce2020reliable}—along with FAB and Square Attack to address challenges such as suboptimal attack steps and objective functions. Besides, adaptive attacks tailor their configurations to evade specific defense mechanisms \cite{athalye2018obfuscated,liu2022practical}.

\textit{Delusive attacks.}  
Unlike testing attacks that deceive well-trained models during inference, delusive attacks \cite{marchant2022hard,liu2022mode} compromise the reliability and availability of models during training. 
Zhe et al. \cite{zhe2024parameter} introduce a parameter matching availability attack that significantly disrupts the training process by perturbing only partial training data.
To protect commercially valuable data in data-sharing scenarios, Zhao et al. \cite{zhao2022guided} propose a double-stream erasable attack, which introduces erasable perturbations into the training data, thereby reducing data availability.
Additionally, the clean-label poisoning availability attack \cite{zhao2022clpa} uses a two-phase GAN and a triplet loss to generate poisoned samples. 
These samples appear realistic to humans but successfully evade the defenses that employ singular value decomposition.

\textit{Ensemble attacks.}  
Cai et al. \cite{cai2023ensemble} demonstrate that a carefully designed ensemble method can realize effective attacks against multiple victim models. 
Specifically, they show that normalizing individual model weights before ensembling is critical for the success of black-box attacks, and further improvement can be achieved by adjusting ensemble weights based on the victim models. To improve the intrinsic transferability of adversarial examples, Chen et al. \cite{chen2023adaptive} propose an adaptive ensemble attack that dynamically adjusts the fusion of model outputs by monitoring the discrepancy ratio of their contributions to the adversarial objective.

\textit{Text attacks.} Jha et al. \cite{jha2023codeattack} present CodeAttack to generate the black-box adversarial code, revealing the vulnerabilities of pre-trained programming language models to code-specific adversarial perturbations. 
Liu et al. \cite{liu2023sspattack} propose SSPAttack, a hard-label textual adversarial method that effectively generates adversarial examples with minimal alterations during the word substitution process.
Likewise, TextCheater \cite{10345721} employs an adaptive semantic preservation optimization technique based on bigram and unigram structures, aiming to reduce word changes required for inducing misclassification.

$\circ$ \textit{AT for defending against various adversarial attacks.}  

\textit{AT for testing attacks.}  
AT can defend against both standard attacks, such as PGD and AutoAttack, and specialized attacks across various tasks. 
For improving the model robustness against standard attacks, 
Dong et al. \cite{dong2023adversarial} replace convolution parameters with randomly sampled mapping parameters, which are varied during the attack and inference processes.
Araujo et al. \cite{araujo2019robust} demonstrate that AT with random noise embedding can resist adversarial attacks constrained by \( \ell_{\infty} \)-norm or \( \ell_2 \)-norm functions. 
Jiang et al. \cite{jiang2023towards} further show that AT can defend against \( \ell_1 \)-bounded attacks by mapping perturbed samples back to the \( \ell_1 \) boundary.
Metzen et al. \cite{metzen2021meta} combine meta-learning with AT to enhance the model robustness against generic patch attacks.

% Shrivastava et al. \cite{shrivastava2017learning} refine adversarial inputs using the discriminator of GANs. 

For defending task-specific adversarial attacks like discrete attacks,  Ivgi et al. \cite{ivgi2021achieving} introduce discrete AT, which leverages symbolic perturbations such as synonym replacements designed for linguistic inputs. 
Gosch et al. \cite{gosch2024adversarial} show that AT effectively mitigates adversarial structural perturbations. 
Ruan et al. \cite{ruan2023towards} propose a viewpoint-invariant AT, where viewpoint transformations are treated as adversarial attacks. 
Specifically, the inner maximization learns a Gaussian mixture distribution to generate diverse adversarial viewpoints, while the outer minimization trains a viewpoint-invariant classifier against the worst-case viewpoint distribution.  
Gao et al. \cite{gao2023effectiveness} investigate the effectiveness of AT in countering backdoor attacks, showing that models trained with spatial adversarial examples can defend against patch-based backdoor attacks.
They also indicate that a hybrid AT strategy can improve robustness against various backdoor attacks. 
Additionally, Zhang et al. \cite{zhang2019limitations} identify a limitation of AT: it primarily adapts to the training distribution and may not effectively improve robustness on samples that deviate from this distribution.

\textit{AT for delusive attacks.}  
To defend against delusive attacks,
Tao et al. \cite{tao2021better} minimize adversarial risk within an $\infty$-Wasserstein ball, which can reduce the reliance of adversarially trained models on non-robust features. They \cite{tao2022can} further reveal that AT with a defense budget of $\epsilon$ is insufficient to guarantee robustness against $\epsilon$-bounded delusive attacks, and propose increasing the perturbation budget in AT. 
Yu et al. \cite{yu2020coded} propose a coded Merkle tree as a hash accumulator to provide a constant-cost protection against data availability attacks.
They \cite{yu2022availability} also show that targeted adversarial perturbations are almost linearly separable and thus can serve as shortcuts for delusive attacks in AT.
% Namely, targeted attacks provide comparable attack intensity with a more efficient generation process than delusive attacks.

\textit{AT for ensemble attacks.}
Conventional AT methods generally concentrate on a single perturbation type, such as adversarial perturbations constrained by a specific norm function (\(\ell_0, \ell_1, \ell_2,\) or \(\ell_\infty\)).
To address this limitation, Tramer et al. \cite{tramer2019adversarial} propose an affine attack that linearly interpolates across various perturbation types. 
Furthermore, adversarial distributional training \cite{dong2020adversarial} designs diverse perturbation distribution functions, from which adversarial examples are sampled. 
Compositional AT \cite{hsiung2023towards} combines multiple perturbation types, including semantic perturbations (\textit{e.g.}, hue, saturation, contrast, brightness) and perturbations in \(\ell_p\)-norm space. Pyramid-AT \cite{herrmann2022pyramid} generates multi-scale adversarial perturbations and aligns the outputs of dropout layers for clean and adversarial examples to enhance AT performance.
Weng et al. \cite{weng2023exploring} focus on non-target classes to solve the dominant bias issue, and employ an AT framework that self-adjusts ensemble model weights to boost the transferability.

\textit{AT for text-related attacks.}  
Unlike continuous visual pixels, text is discrete, prompting researchers to apply adversarial perturbations to text embeddings \cite{wang2019improving}. 
Specifically, for graph networks like DeepWalk \cite{perozzi2014deepwalk}, Dai et al. \cite{dai2019adversarial} generate negative graphs based on node dependencies and modify graph embeddings with adversarial attacks. To reduce the discrepancy between  generated adversarial text examples and the text encoder, Zhao et al. \cite{zhao2023generative} employ a discriminator to refine text embedding layers and a generator to yield adversarial examples for masked texts.

$\circ$ \textit{Attack Intensity Adjustment.}
Under the same attack, adjusting the attack intensity can greatly affect AT performance.

\textit{AT with small perturbation degree.}
Liu et al. \cite{liu2020loss} observe that large adversarial budgets during training can hinder the escape from suboptimal perturbation initialization. 
To address this, they propose a periodic adversarial scheduling strategy that dynamically adjusts the perturbation budget, similar to learning rate warmup. 
To enhance the prediction performance of adversarially trained models on clean samples, Kim et al. \cite{kim2021understanding} verify multiple perturbation budgets in each attack iteration, selecting the smallest budget that successfully misleads the target model.
Yang et al. \cite{yang2022one} reduce the perturbation budget when adversarial examples excessively cross the decision boundary. 
Friendly AT \cite{zhang2020attacks} terminates the attack iteration early to prevent generating overly strong perturbations.
Furthermore, Wang et al. \cite{wang2021convergence} indicate that using high-quality adversarial examples early in AT can degrade adversarial robustness.
Shaeiri et al. \cite{shaeiri2020towards} first train models with small  perturbations, followed by fine-tuning models on larger ones. 
Curriculum AT \cite{shi2024closer} gradually increases the hardness of adversarial examples, which are sampled from time-varying distributions using a series prediction framework.
% and derive generalization error bounds of curriculum AT.
Hybrid AT \cite{li2020towards} starts with FGSM and shifts to stronger attacks like PGD to mitigate catastrophic overfitting.
% , a phenomenon where model robustness drops to near zero during training.

\textit{AT with large perturbation degree.}
Conversely, Gong et al. \cite{gong2021maxup} propose MaxUp to enhance adversarial generalization by performing the outer minimization on the worst-case adversarial examples. 
Specifically, they generate multiple adversarial examples for each input and select the most aggressive one. 
Xu et al. \cite{xu2022a2} propose a parameterized automatic attacker to search for optimal adversarial perturbations. 
Liang et al. \cite{liang2022efficient} estimate the worst-case reward for the reinforcement learning policy using \( \ell_p \)-bounded attacks.
De et al. \cite{de2022make} show that using stronger adversarial perturbations, such as removing the clipping operation, can mitigate catastrophic overfitting.

\textit{AT with learnable perturbation degree.}
Xiong et al. \cite{xiong2020improved} design an optimizer with recurrent neural networks to predict adaptive update directions and hyperparameters for the inner maximization. Jia et al. \cite{jia2022adversarial} introduce a learnable attack strategy that dynamically adjusts hyperparameters such as the number of attack iterations, the maximum perturbation size, and the attack stride.
Ye et al. \cite{ye2021annealing} introduce the Amata annealing mechanism, which gradually increases the number of attack steps in inner maximization.

\textit{Adversarial ratio adjustment.}
Yoo et al. \cite{yoo2021towards} generate adversarial examples at each training epoch and randomly replace a certain percentage of clean examples with these adversarial examples. 
Dolatabadi et al. \cite{dolatabadi2022} also argue that AT does not require calculating adversarial perturbations for every input. They employ a greedy algorithm to select a training subset for perturbation, where the weighted gradients of this subset can approximate the gradient statistics of the entire training dataset.

\subsection{Model Perspective}
% Based on the paradigm in Eq. (\ref{eqinint}), it is evident that A
AT does not prescribe a specific network architecture.
This subsection mainly describes AT methods that explore various network structures and components.

\underline{\textit{1) Network Structures:}}

$\circ$ \textit{Vision-based.}
Salman et al. \cite{salman2020adversarially} demonstrate that CNNs and Transformers exhibit comparable adversarial robustness under identical conditions. 
Singh et al. \cite{singh2024revisiting} perform comparative experiments between ConvNeXts \cite{liu2022convnet} and Vision Transformers (ViTs) \cite{touvron2021training} on the large-scale ImageNet dataset \cite{krizhevsky2012imagenet}. They observe that replacing the PatchStem with ConvStem \cite{xiao2021early} significantly improves the adversarial robustness of both networks.
Additionally, the ConvNeXt with ConvStem performs the highest robustness against $\ell_\infty$-bounded threats, while the ViT with ConvStem excels in robust generalization to unseen models. Mo et al. \cite{mo2022adversarial} highlight the critical role of pre-training for ViTs in AT. They also show that employing the SGD optimizer and extending the number of training epochs can further enhance AT performance.

\begin{table*}[ht]
    \centering
    \caption{Performance comparison of different models on ImageNet-1K. ViT-H/14 and ViT-g/14 additionally incorporate the training data from DataComp-1B \cite{gadre2024datacomp}. \(\ell_\infty, \ell_2, \ell_1\) are adversarial perturbation metrics.}
    \label{tab:model_comparison}
    % \resizebox{0.9\textwidth}{!}{
    \begin{tabular}{l|cccc|cccc}
    \toprule[1pt]
    \textbf{Model} & \textbf{Pre-trained} & \textbf{Epochs} & \textbf{PGD Steps}&\textbf{Params (M)} & \textbf{Clean} & \textbf{$\ell_\infty$} & \textbf{$\ell_2$} & \textbf{$\ell_1$} \\
    \midrule[0.5pt]
    ResNet-50 \cite{bai2021transformers} &\ding{55} &100 &1 & 25&67.4 &35.5 &18.2& 3.90\\
    % \hline
    Wide-ResNet-50-2 \cite{salman2020adversarially} &\ding{55}&100&3&68.9 & 68.8 &38.1 &22.1 &4.48\\
    % \hline
    ViT-S \cite{bai2021transformers}&\ding{55} &100 &1 &22.1 &66.6 &36.6 &41.4 &21.8\\
    % \hline
    XCiT-S12 \cite{debenedetti2023light} &\ding{55} &110 &1 &26&72.3 &41.8 &46.2 &22.7\\
    % \hline
    ConvNeXt-T \cite{debenedetti2023light}&\ding{55}&110&1& 28.6 &71.6 &44.4 &45.3 &21.8\\
    % \hline
    RobArch-L \cite{peng2023robarch} &\ding{55} &100&3 & 104&73.5 & 48.9 & 39.5 & 14.7 \\
    % \hline
    ViT-B/16 \cite{rebuffi2022revisiting} &\ding{55}&300&2&87 & 76.6 & 53.5 & - & - \\
    % \hline
    ConvNeXT-B \cite{liu2024comprehensive}&\ding{55}&50&2& 89  & 76.0 & 55.8 & 44.7 & 21.2 \\
    % \hline
    Swin-B \cite{liu2024comprehensive} &\ding{55}&300&3&88  & 76.2 & 56.2 & 47.9 & 23.9 \\
    % \hline
    ConvNeXT-B+ConvStem \cite{singh2024revisiting}&\ding{51}&50&2 &89   & 75.2 & 56.3 & 49.4 & 23.6 \\
    % \hline
    % ConvNeXT-L+ConvStem \cite{singh2024revisiting}&Y&250&2 &198  & 77.0 & 57.7 & 47.0 & 22.2 \\\hline
    ConvNeXT-L+ConvStem \cite{singh2024revisiting}&\ding{51}&250&3& 198  & 78.2 & 59.4 & 56.2 & 33.8 \\
    % \hline
    ConvNeXT-L \cite{liu2024comprehensive}&\ding{55}&-&3& 198 & 78.0 & 58.5 & - & - \\
    % \hline
    Swin-L \cite{liu2024comprehensive}&\ding{55}&-&3& 197  & 78.9 & 59.6 & - & - \\
    % \hline
    ViT-H/14 \cite{wang2024revisiting} &\ding{55}&200&2/3& 304 & 83.9 & 69.8 & 69.8 & 46.0 \\
    % \hline
    ViT-g/14 \cite{wang2024revisiting} &\ding{55}&200&2/3& 1013 & 83.9 & 71.0 & 70.4 & 46.7 \\
    \bottomrule[1pt]
    \end{tabular}
    % }
    \vspace{-2mm}
\end{table*}

$\circ$ \textit{Language-based.}
Feng et al. \cite{feng2019graph} introduce Graph AT to improve the robustness and generalization of GNNs, which leverages sample relationships to construct adversarial perturbations.
Morris et al. \cite{morris2020textattack} develop a natural language processing framework that incorporates text adversarial attacks, data augmentation, and AT. 
Ziegler et al. \cite{ziegler2022adversarial} indicate that AT techniques mitigate the risk of catastrophic failures in generative language models, leading to high-quality completions for given prompts. 
In text classification, Pan et al. \cite{pan2022improved} enhance transformer-based encoders by generating adversarial perturbations for the word embedding matrix and performing contrastive learning on both clean and adversarial examples.
Chen et al. \cite{chen2022adversarial} propose a variational word mask network to interpret the robustness brought by AT.
They find that AT helps identify important words and resolve feature mismatches between clean and adversarial example pairs.

$\circ$ \textit{Cross-modal.}
Gan et al. \cite{gan2020large} introduce AT into the large-scale visual and language representation learning task, implementing AT individually in the embedding space of each modality. 
Li et al. \cite{li2024one} improve the robustness of vision-language models through adversarial textual prompt learning.
Wu et al. \cite{wu2020augmented} propose a cross-modal retrieval method that aligns data from different modalities through augmented AT. They integrate semantic information from the label space into the AT process by sampling semantically relevant and irrelevant source-target pairs.
CMLA \cite{li2019cross} introduces similarity regularization constraints for both inter-modal and intra-modal during adversarial learning, increasing the difficulty of cross-modal adversarial example generation.
Sheng et al. \cite{sheng2021cross} present a self-supervised adversarial learning approach for lip reading, utilizing an adversarial dual contrastive learning technique to learn speech-relevant visual representations.
% Tang et al. \cite{tang2019adversarial} demonstrate that AT can improve the adversarial robustness of multimedia recommendation systems. 

$\circ$ \textit{Multi-model.} 
Vivek et al. \cite{vivek2018gray} point out that current AT methods focus on evaluating optimal models, overlooking the behavior of intermediate models. To address this, they propose periodically saving intermediate models during training and assessing the adversarial robustness of each saved model. Rebuffi et al. \cite{rebuffi2022revisiting} demonstrate that separating trainable parameters for classifying clean and adversarial examples can significantly reduce the performance trade-off between the two. Liu et al. \cite{liu2022mutual} introduce a collaborative AT approach, in which multiple models are jointly trained to share adversarial knowledge, thereby enhancing the adversarial robustness of each model.
Additional multi-model AT methods, such as robust fine-tuning and robust distillation techniques, are discussed in the Loss Function section.

Table \ref{tab:model_comparison} shows the performance comparison of different models on the Imagenet dataset.

\underline{\textit{2) Network Components:}}

$\circ$ \textit{Activation functions.}
Xie et al. \cite{xie2020smooth} enhance AT performance by replacing the ReLU activation with smoother alternatives such as Softplus \cite{dugas2000incorporating}, SiLU \cite{ramachandran2017searching}, GELU \cite{hendrycks2016gaussian}, and ELU \cite{clevert2015fast}. 
Similarly, Singla et al. \cite{singla2021low} argue that activation functions like SiLU and LeakyReLU improve model robustness by improving gradient stability. They also demonstrate the effectiveness of these activation functions in mitigating catastrophic overfitting.
The formulations of popular activation functions in AT are as follows:
\begin{itemize}
    \item[-] Rectified Linear Unit (RELU):
\[ \text{ReLU}(x) = \max(0, x). \]
\item[-] Leaky Rectified Linear Unit (LeakyReLU):
\[ \text{LeakyReLU}(x) = \max(\alpha x, x). \]
\item[-] Smooth ReLU (Softplus),
\[ \text{Softplus}(x) = \log(1 + \text{exp}(x)).\]
\item[-] Sigmoid Linear Unit (SiLU):
\[ \text{SiLU}(x) = \frac{x}{1 + \text{exp}(-x)}. \]
\item[-] Gaussian Error Linear Unit (GELU):
\[ \text{GELU}(x) = x \cdot \frac{1}{2} \left[ 1 + \text{erf}\left( \frac{x}{\sqrt{2}} \right) \right], \]
\[ \text{erf}(x) = \frac{2}{\sqrt{\pi}} \int_{0}^{x} \text{exp}(-t^2) dt. \]
\item[-] Exponential Linear Unit (ELU):
\[ \text{ELU}(x) = \begin{cases}
x & \text{if } x \geq 0 \\
\beta\cdot (\text{exp}(x) - 1) & \text{if } x < 0
\end{cases}
\]
\end{itemize}
where the default values of the hyperparameters \(\alpha\) and \(\beta\) are set to 0.01 and 1, respectively.

$\circ$ \textit{Batch Normalization (BN).} Based on the clean-adversarial domain hypothesis, Xie et al. \cite{xie2019intriguing} demonstrate that using separate BN layers for clean and adversarial examples effectively reduces the performance trade-off. Addepalli et al. \cite{addepalli2022efficient} extend this approach by introducing a joint AT method with diversity augmentation, also utilizing distinct BN layers for clean and adversarial examples. 
Liu et al. \cite{liu2023twins} design a dual-branch architecture with distinct BN layers: the first BN layer leverages statistics from the pre-training dataset, while the second BN layer utilizes statistics from each minibatch of the target dataset. Adversarial examples are crafted for the second branch, and both branches are optimized jointly.
Wang et al. \cite{wang2022removing} attribute the performance degradation of adversarially trained models on clean samples to the difficulty that BN layers face in handling mixed data distributions. To address this, they propose a normalizer-free robust training method.

$\circ$ \textit{Dropout strategy.}
% Considering that learning from more aggressive adversarial examples can improve AT performance, 
Wu et al. \cite{wu2022towards} propose an attention-guided dropout strategy that selectively masks partial network embeddings during training. Vivek et al. \cite{vivek2020single} introduce a dropout scheduling technique, where dropout layers are applied after each network layer, with the dropout rate gradually decreasing as training progresses.

$\circ$ \textit{Other components.}
Huang et al. \cite{huang2023revisiting} explore the impact of various architectural elements in residual networks, such as topology, network depth, and channel width, on adversarial robustness. They propose RobustResBlock and RobustScaling to enhance network robustness by adjusting block structures and scaling strategies, respectively.

\subsection{Training Configurations Perspective}\label{III-C-1}
This subsection further investigates the influence of various training configurations, such as loss functions, labels, and optimization algorithms, on AT performance.

\underline{\textit{1) Loss Functions:}}

$\circ$ \textit{Loss functions in conventional AT.}
Conventional AT methods \cite{zhang2019theoretically,wang2019bilateral} such as PGD-AT, are known for their stability and effectiveness in improving model robustness, requiring multiple gradient descent steps to yield adversarial data.

Specifically, the inner maximization often employs the cross-entropy loss \(\ell_\text{ce}\).
Regarding the outer minimization, TRADES \cite{zhang2019theoretically} incorporates the KL divergence,
\[ \min_\theta\ell_{\text{TRADES}} = \ell_\text{ce}(x, y; \theta) + \beta\cdot {\text{KL}}(f({x} + \delta; \theta)\|f({x}; \theta)), \]
where \(x\) means clean samples. \(\beta\) controls the performance trade-off between clean and adversarial examples.
\( f(\cdot;\theta)\) denotes the trained model with weights \(\theta\). 
% Instead of using KL divergence, 
Nuclear AT \cite{sriramanan2021towards} utilizes the nuclear norm function \(\|\cdot\|_*\) to constrain the prediction differences between the two types of examples,
\[\min_{\theta} [\ell_\text{ce}(x, y;\theta) + \|f(x+\delta;\theta) - f(x;\theta)\|_*],\]
Cui et al. \cite{cui2021learnable} propose a learnable boundary-guided AT, which employs cross-entropy and regularization losses to supervise the model weights for classifying clean and adversarial examples, respectively,
\[\min_{\theta_\text{clean}, \theta_\text{AT}} [\ell_\text{ce}(x, y;\theta_\text{clean}) + \|f(x+\delta;\theta_{AT}) - f(x;\theta_\text{clean})\|_p].\]
% where \(\|\cdot\|_p\) refers to the \(p\)-norm function.
RLAT \cite{kireev2022effectiveness} replaces the KL divergence in \(\ell_{\text{TRADES}}\) with LPIPS \cite{zhang2018unreasonable} to perceive image patch similarity,
\[ \text{LPIPS}(x, x+\delta) = \sum_{l=1}^L \alpha_l\cdot\|f_l(x+\delta) - f_l(x)\|_2^2, \]
where \(\|f_l(x+\delta) - f_l(x)\|_2^2\) denotes the squared Euclidean distance between the extracted features for \(x+\delta\) and \(x\) at the layer \(l\). \(\alpha_l\) represents the layer weighting factor.

The above methods often regularize the model predictions for \(x+\delta\) and \(x\), 
which can affect AT performance when clean examples are misclassified.
To tackle this, Dong et al. \cite{dong2023enemy} craft adversarial perturbations $\delta_\text{clean}$ for clean samples, ensuring all crafted samples \(x+\delta_\text{clean}\) can be correctly classified.
Then, they regularize the predictions for \(x+\delta\) and \(x+\delta_\text{clean}\),
\[\min_\theta [\ell_\text{ce}(x, y; \theta) +\|f(x+\delta;\theta) - f(x+\delta_\text{clean};\theta)\|_p].\]
Fan et al. \cite{fan2021adversarial} combine adversarial and abstraction losses to achieve provably robust AT.

In addition to using \(\ell_{ce}\) in the inner maximization, Zhang et al. \cite{zhang2019defense} maximize the feature distribution distance between clean and adversarial examples based on optimal transport metrics.
% , which is an unsupervised approach.
Zou et al. \cite{zou2021provable} leverage AT to develop robust classifiers against unknown label noise. They investigate various loss functions in the inner maximization, such as binary cross-entropy and non-convex S-shaped losses, and indicate that choosing an appropriate loss function can greatly improve AT performance.
Addepalli et al. \cite{addepalli2022scaling} observe that large adversarial perturbations may alter the true categories of inputs. To address this, they introduce distinct constraints for oracle-invariant and oracle-sensitive samples. These samples are divided based on whether crafted perturbations modify human cognition. 
For oracle-invariant samples, the inner maximization is formulated as 
\[\max_\delta[\ell_\text{ce}(x+\delta,y;\theta)-\lambda\cdot\text{LPIPS}(x, x+\delta)],\] 
while the outer minimization utilizes $\ell_\text{TRADES}$. For oracle-sensitive samples, the LPIPS term is excluded from the inner maximization, and the outer minimization is expressed as
\[\min_\theta [\ell_\text{hybrid} + \beta\cdot {\text{KL}}(f({x} + \delta; \theta)\|f({x}; \theta))],\]
\[\ell_\text{hybrid} = \alpha \cdot \ell_\text{ce}(x, y;\theta) + (1 - \alpha) \cdot \ell_\text{ce}(x+\delta, y;\theta).\]

$\circ$ \textit{Loss functions in fast AT.}
Fast AT, which employs single-step adversarial attacks like FGSM in the inner maximization, is time-efficient but prone to catastrophic overfitting \cite{zhao2024catastrophic}.

Regarding the outer minimization, FGSM-AT \cite{goodfellow2014explaining} directly utilizes  \(\ell_\text{hybrid}\).
% , which directly incorporates clean data \(x\) and adversarial data \(x + \delta\) during training.
Li et al. \cite{li2018learning} further introduce a supplementary regularization for \(\ell_\text{hybrid}\) to improve adversarial robustness,
\[\sum_i\beta_i\cdot\sum_{j} \|f_{i,j}(x; \theta) - f_{i,j}(x+\delta; \theta)\|_p,\]
where \(j\) and \(i\) are indices of the feature channels and layers, respectively. 
% \( \beta_i \) is hyperparameters. 
The prior-guided initialization method \cite{jia2022prior} introduces a regularization constraint for adversarial examples \(x+\delta\) and initially perturbed samples $x+\delta_0$,
\[
\beta\cdot\|f(x+\delta;\theta) - f(x+\delta_0;\theta)\|_2^2,
\]
where \( \delta_0 \) denotes uniformly sampled random perturbations, \( \delta_0\in\text{U}(- \epsilon, \epsilon) \).
\(\epsilon\) represents the maximum perturbation value.
The default value of $\beta$ is set to 10.
Besides, Zhao et al. \cite{zhao2023fast} demonstrate the adversarial vulnerability of exiting AT methods on large perturbation budgets.
To address this, they select the maximum activation layer based on the feature activation differences between \(x + \delta\) and \(x + \delta_0\), and restrict the feature differences at the selected layer \cite{zhao2024catastrophic}:
\[
\beta\cdot\|f_\text{act-max}(x + \delta; \theta) - f_\text{act-max}(x + \delta_0; \theta)\|_2^2,
\]
where \(f_\text{act-max}\) denotes the maximum feature activation layer. The default value of $\beta$ is set to 300.
Additionally, they \cite{zhao2023fast} attribute catastrophic overfitting to convergence anomalies in specific samples and mitigate it by restricting the convergence speed of these outliers,
\begin{equation*}
\begin{aligned}
        |\ell_\text{ce}(x+\delta,y;\theta) - \mu_\text{pre}^\prime| + &|\ell_\text{ce}(x,y;\theta) - \mu_\text{pre}|,\\ &s.t.\, |\ell_\text{ce}(x,y;\theta) - \mu_\text{pre}|>\gamma,
\end{aligned}
\end{equation*}
or restricting the convergence speed of abnormal batches, 
\begin{equation*}
\begin{aligned}
        |\mu_\text{batch}^\prime - \mu_\text{pre}^\prime| + |\mu_\text{batch} - \mu_\text{pre}|,\, s.t.\, |\mu_\text{batch} - \mu_\text{pre}|>\gamma,
\end{aligned}
\end{equation*}
where \(\mu_\text{pre}^\prime = \mathbb{E}_{(x,y)\sim \mathcal{D}}\ell_\text{ce}(x+\delta,y;\theta_\text{pre})\) and \(\mu_\text{pre} = \mathbb{E}_{(x,y)\sim \mathcal{D}}\ell_\text{ce}(x,y;\theta_\text{pre})\).
\(\theta_\text{pre}\) represents the weights in the previous epoch.
\(|\cdot|\) calculates the absolute value.
\(\gamma\) is the convergence stride used for selecting abnormal data.

Unlike these methods that restrict the deviation between predicted features, GradAlign \cite{andriushchenko2020understanding} aligns the back-propagated gradients for perturbed examples with those for clean samples,
\[
 1 - \cos \left( \nabla_{x+\delta_0} \ell_\text{ce}(x+\delta_0, y; \theta), \nabla_x \ell_\text{ce}(x, y; \theta) \right).
\]

The experimental results of conventional and fast AT techniques with the budget of \(\epsilon = 8/256\) are shown in Table \ref{tab:performance}.
% The supplement displays the comparison of different methods on various perturbation budgets.
Additionally, Table \ref{tab:adversarial_performance} displays the comparison of different methods on various perturbation budgets.

\begin{table*}[ht]
    \centering
    \caption{Performance comparison of AT methods on CIFAR10 \cite{krizhevsky2009learning} and ResNet \cite{he2016deep}. `Time' refers to the running minutes on a 4090 GPU. 
The total training duration is 110 epochs, with the learning rate decaying by a factor of 0.1 at the 100th and 105th epochs.}
    \label{tab:performance}
    % \resizebox{0.93\textwidth}{!}{
    \begin{tabular}{l|ccccccccccc}
    \toprule[1pt]
    \textbf{Method (\(\epsilon= 8/256\) )} && \textbf{Clean Acc} & \textbf{FGSM} & \textbf{PGD10} & \textbf{PGD20} & \textbf{PGD50} & \textbf{C\&W} & \textbf{APGD} & \textbf{Square} & \textbf{AutoAttack} & \textbf{Time} \\
    \hline
    ORI \cite{he2016deep}&-&94.40&2.13&0&0&0&0&0&-&0&24\\
    \hline
    \multirow{2}{*}{PGD-AT \cite{rice2020overfitting}}& Best& 82.57 & 63.93 & 53.19 & 52.42 & 52.21 & 48.01 & 51.22 & 55.70 & 48.77 & \multirow{2}{*}{199} \\
    & Last & 82.99 & 64.20 & 53.05 & 52.14 & 51.96 & 47.51 & 50.66 & 55.23 & 48.23 &  \\
    \hline
    \multirow{2}{*}{TRADES \cite{zhang2019theoretically}}& Best & 82.03 & 64.28 & 54.06 & 53.35 & 53.16 & 46.22 & 50.66 & 55.83 & 49.47 & \multirow{2}{*}{241} \\
    & Last & 81.80 & 63.57 & 53.89 & 53.23 & 53.05 & 46.31 & 50.39 & 55.54 & 49.56 &  \\
    \hline
    \multirow{2}{*}{Free-AT \cite{shafahi2019adversarial}} &Best & 80.64 & 57.62 & 44.10 & 42.99 & 42.73 & 39.88 & 42.83 & 48.82 & 41.17 & \multirow{2}{*}{140} \\
     &Last & 79.88 & 55.4 & 42.00 & 40.49 & 40.18 & 38.83 & 40.37 & 46.03 & 40.34 & \\
    \hline
    LBGAT \cite{cui2021learnable}&Best&81.98&-&-&57.78&-&-&-&-&53.14&-\\
    \hline
    OAAT \cite{addepalli2022scaling}&Best&80.24&78.54&-&-&-&-&-&-&50.88&-\\
    \hline
    \multirow{2}{*}{GAT \cite{sriramanan2020guided}} &Best & 79.79 & 54.18 & 53.55 & 53.42 & 49.04 &- & 47.53 &- &-&\multirow{2}{*}{87}\\
    &Last & 80.41 & 53.29 & 52.06 & 51.76 & 49.07 &- & 46.56 &- &-&\\
    \hline
    \multirow{2}{*}{NuAT \cite{sriramanan2021towards}} &Best & 81.58 & 53.96 & 52.90 & 52.61 & 51.30 &- & 49.09 &- &-&\multirow{2}{*}{79}\\
    &Last & 81.38 & 53.52 & 52.65 & 52.48 & 50.63 &- & 48.70 &-&-&  \\
    \hline
    UIAT\cite{dong2023enemy}&Best&85.01&-&54.63&-&-&-&-&-&49.11&-\\
    \hline
    \multirow{2}{*}{GradAlign \cite{andriushchenko2020understanding}} &Best & 81.27 & 59.52 & 45.49 & 43.84 & 43.52 & 42.84 & 43.15 & 40.96 & 39.29 & \multirow{2}{*}{132} \\
    &Last & 86.40 & 0 & 0 & 0 & 0 & 0 & 0 & 0 & 0 & \\
    \hline
    \multirow{2}{*}{FGSM-RS \cite{wong2020fast}}& Best & 72.95 & 52.64 & 41.40 & 40.57 & 40.40 & 36.83 & 39.95 & 44.75 & 37.67 & \multirow{2}{*}{40} \\
    & Last & 84.24 & 0 & 0 & 0 & 0 & 0 & 0 & 0 & 0 &\\
    \hline
    \multirow{2}{*}{FGSM-CKPT \cite{kim2021understanding}} &Best& 90.29&-& 41.96& 39.84& 39.15 &41.13&-&-& 37.15& \multirow{2}{*}{58} \\
    &Last &90.29 &-&41.96 &39.84& 39.15 &41.13&-&- &37.15&\\ 
    \hline
    N-FGSM \cite{de2022make} &Best&80.48	&-&-&-&47.91&-&-&-&44.81&-\\
    \hline
    \multirow{2}{*}{FGSM-EP \cite{jia2022prior}} &Best & 79.28 & 65.04 & 54.34 & 53.52 & 53.54 & 45.84 & 47.85 & 52.42 & 45.66 & \multirow{2}{*}{57} \\
    &Last & 80.91 & 73.83 & 40.89 & 37.76 & 36.05 & 38.46 & 28.41 & 45.29 & 25.16 & \\
    \hline
    \multirow{2}{*}{FGSM-MEP \cite{jia2022prior}} &Best & 81.72 & 64.71 & 55.13 & 54.45 & 54.29 & 47.05 & 50.39 & 55.47 & 48.23 & \multirow{2}{*}{57} \\
    &Last & 81.72 & 64.71 & 55.13 & 54.45 & 54.29 & 47.05 & 50.39 & 55.47 & 48.23 &  \\
    \hline
    \multirow{2}{*}{FGSM-PCO \cite{wang2024preventing}} &Best & 82.05 & 65.53 & 56.32 & 55.66 & 55.67 & 47.12 & 50.05 & 55.59 & 48.04 & \multirow{2}{*}{60} \\
    &Last & 82.05 & 65.53 & 56.32 & 55.66 & 55.67 & 47.12 & 50.05 & 55.59 & 48.04 & \\
    % \hline
    \bottomrule[1pt]
    \end{tabular}
    % }
    % \vspace{-2mm}
\end{table*}

$\circ$ \textit{Loss functions in federated AT.}
Federated AT \cite{zizzo2020fat} computes gradients for each local client \(i\) individually across \( N \) local samples \(\{{x}_{j}, y_{j}\}_{j=1}^{N}\) as follows:
\[
\sum_{j=1}^{K} \nabla_{\theta_i} \ell_\text{ce} \left( {x}^{\prime}_{j}, y_{j}; \theta_i \right) + \sum_{j=K+1}^{N} \nabla_{\theta_i} \ell_\text{ce} \left( {x}_{j}, y_{j}; \theta_i \right),
\]
where \( K \) determines the proportion of adversarial examples \(x^\prime\), which are typically generated by FGSM or PGD.
The gradients from various local models are then aggregated by a central server to construct a global model.

% Ensuring security in federated AT poses significant challenges, particularly when . 
In federated AT, some local servers may behave maliciously.
To detect such stealthy attacks, Zizzo et al. \cite{zizzo2021certified} introduce certifiable federated AT based on the abstract interpretation techniques. 
Given a network \( f \), they verify that the predicted label for any input within a \( p \)-norm ball \( B_p^\epsilon \) of radius \( \epsilon \), centered around \( x \), remains consistent with the true class \( y \):
\[
\text{arg max} \ f(x';\theta) = y, \ \forall x' \in B_p^\epsilon(x) = \{x' = x + \delta \ | \ \|\delta\|_p \leq \epsilon \}.
\]
Practically, they utilize an abstract element \( \hat{x} \) to represent the \( p \)-norm ball of \( x \):
\[
\hat{x} = \eta_0 + \sum_{i=1}^{M} \eta_i\cdot\epsilon_i.
\]
Here, \( \eta_0 \) denotes the central coefficient and \( \eta_i \) is the deviation from the center, with an associated error symbol \( \epsilon_i \), where \( \epsilon_i \) can be either +1 or -1. 
\(M\) means the number of deviation points.
Federated AT also faces various challenges under non-independent and identically distributed (non-IID) conditions.
Specifically, to address training instability and decreased clean classification accuracy, Chen et al. \cite{chen2022calfat} propose calibrated federated AT that minimizes the following objective:
\[
\min_{\theta_i} \frac{1}{n_i} \sum_{j=1}^{n_i} -\log \sigma(f({x}_{ij}^\prime;\theta_i) + \log \pi_i),
\]
where \(\sigma\) is the Softmax function, \( x_{ij} \) is the \(j\)-th example of the \(i\)-th client, and \(\pi_i\) is a constant computed based on the local dataset \(D_i\). 
\(n_i\) denotes the number of examples in \(D_i\).
\( {x}_{ij}^\prime \) represents the adversarial example of \( x_{ij} \), generated by
% Adversarial examples are generated by maximizing the calibrated Kullback–Leibler divergence loss:
\[
\arg \max_{x_{ij}^\prime} [- \sigma(f(x_{ij};\theta_i) + \log \pi_i) \log \sigma(f(x_{ij}^\prime;\theta_i) + \log \pi_i)].
\]
To resolve bias and fairness issues, Zhang et al. \cite{zhang2023delving} assign higher weights to samples near the decision boundary and ensure consistency between local model predictions for adversarial examples and global model predictions for clean inputs. Given a local model \( f_{\text{loc}} \) and a global model \( f_{\text{glo}} \), the local objective is formulated as:
\[
\min_{\theta_\text{loc}} [\ell_{\text{ce}}(\rho \cdot f_{\text{loc}}(x^\prime;\theta_\text{loc}), y) + \beta \cdot \text{KL}(f_{\text{loc}}(x^\prime;\theta_\text{loc})\| f_{\text{glo}}(x;\theta_\text{glo}))].
\]
For a mini-batch of samples \( \{x_i, y_i\}_{i=1}^m \), the parameter list \( \rho_i\in\rho \) is expressed as follows,
\[
\rho_i \leftarrow 1 - \left\{\frac{d_i}{\sum_{j=1}^m d_j}\right\},
\]
where \(d_i\) represents the minimum attack step to deceive the \(i\)-th local model.

$\circ$ \textit{Loss functions in virtual AT (VAT).}
VAT is a semi-supervised AT method. 
For unsupervised samples, Miyato et al. \cite{miyato2018virtual} utilize their predictions from a pre-trained model as virtual labels, minimizing the following objective
\begin{equation}\label{vat}
\min_{\theta}\mathbb{E}_{(x,y)\sim D_{L}} [-\log p(y|x; \theta)] + \mathbb{E}_{x\sim D_{U}} [ \ell_\text{virtual}(\theta, \hat{\theta}, x, \epsilon)],
\end{equation}
where \( D_{L} \) and \( D_{U} \) represent labeled and unlabeled datasets, respectively. \(\ell_\text{virtual}\) is formulated as:
\[
\ell_\text{virtual}(\theta, \hat{\theta}, {x}, \epsilon) = - \sum_{k=1}^{K} p(k|x; \hat{\theta}) \log p(k|x + \delta; \theta),
\]
where \(K\) is the number of categories. 
\( \hat{\theta} \) denotes the current estimate of \( \theta \).
\( \delta = \arg\max_{\delta, \|\delta\|\leq \epsilon} \text{KL} ( p(\cdot | x; \hat{\theta}) \| p(\cdot|x + \delta; \hat{\theta}) ) \).

For vision-based VAT, Kim et al. \cite{kim2019understanding} employ Bad-GAN \cite{dai2017good} to generate high-quality bad examples and introduce a supplementary regularization term for Eq. (\ref{vat}), 
\[
\lambda \cdot \left[ \mathbb{E}_{x \sim D_{U}} \ell_{\text{u}}(x; \theta) + \mathbb{E}_{x \sim D_{\text{GAN}}} \ell_{\text{GAN}}(x; \theta)\right],
\]
where \( D_{\text{GAN}} \) denotes the set of generated adversarial samples. 
\(\ell_{\text{u}}\) and \(\ell_{\text{GAN}}\) are loss functions for unlabeled and generated examples, respectively.
The former is defined as:
{\footnotesize
\begin{equation*}
- \sum_{k=1}^{K} \left[ \frac{\exp(f_k(x; \theta))}{1 + \sum_{k'=1}^{K} \exp(f_{k'}(x; \theta))} \log \frac{\exp(f_k(x; \theta))}{1 + \sum_{k'=1}^{K} \exp(f_{k'}(x; \theta))} \right],        
\end{equation*}
}
and the latter is expressed as:
\[
\log ({1 + \sum_{k=1}^{K} \exp(f_k(x; \theta))}).
\]
Osada et al. \cite{osada2020regularization} introduce a latent VAT technique, where adversarial perturbations are injected into the latent space based on generative models such as variational autoencoders.
\[
\delta = \arg \max_{\delta} \text{KL}(f(x; \theta) \| f(x^\prime; \theta)), \, s.t.\, \|\delta\|_2 \leq \epsilon,
\]
\[
x^\prime = \text{Dec}(\text{Enc}(x) + \delta).
\]

Regarding language-based VAT, Chen et al. \cite{chen2020seqvat} propose SeqVAT, which effectively integrates VAT with Conditional Random Fields (CRFs) for constructing sequence labeling models. The adversarial loss in SeqVAT is formulated as:
\[
\min_\theta \text{KL}\left(f(S; w, c, {\theta}) \parallel f(S; w + \delta_w, c + \delta_c, {\theta})\right),
\]
where \( f(S; w, c, {\theta}) = (p_1, p_2, \ldots, p_k, 1 - \sum_{i=1}^{k} p_i) \) represents the estimated probability distribution over the \( k \)-best label sequences for the given input embeddings \( w \) and \( c \). \( p_i = p_{\text{crf}}(s_i; w, c, {\theta}) \) denotes the computed probability by the CRF model for the \( i \)-th label sequence \( s_i \in S\), with \( i \in [1, k] \). \( \delta_w \) and \( \delta_c \) are adversarial perturbations applied to \( w \) and \( c \), respectively.
Li et al. \cite{li2021token} propose a token-aware VAT method for the language understanding task, generating token-level accumulated perturbations for initialization and using token-level normalization balls to constrain these perturbations.
Lee et al. \cite{lee2022context} present a context-based VAT technique for the text classification task.
This technique preserves contextual semantics and prevents overfitting to noisy labels based on the context-driven label smoothing loss.

% expressed as
% \[
 % \min_\theta -\frac{1}{N} \sum_{i=1}^{N} \log P(f(x;\theta), y^\prime;\theta_{sm}) + \frac{\lambda}{N} \sum_{i=1}^{N} \text{CLS}(x, \theta_{sm}),
% \]
% \[
% P(c, y^\prime; \theta_{\text{sm}}) = \frac{\text{exp}(c^\top w_{y^\prime})}{\sum_{k=1}^K \text{exp}({c^\top w_k})},
% \]
% where \( y^\prime \) is a potential flipped label of \( y \). 
% \(\theta_{sm}\) is the trainable parameter \(w_k\) of the softmax layer.
% CLS denotes a context-driven label smoothing constraint,
% \[
% \text{CLS}(x, \theta_{sm}) = \text{KL}\left(P(x, \cdot;\theta_{sm}) \parallel P(x + \epsilon \cdot \frac{g}{\|g\|_2}, \cdot; \theta_{sm})\right),
% \]
% with the gradient \( g \) calculated by
% \[
% g = \nabla_\delta \text{KL}\left(f(x ; \theta) \parallel f(x + \delta ; \theta)\right).
% \]

\begin{table*}[t]
    \centering
    \tabcolsep = 0.07cm
    \caption{Performance comparison of AT methods on various perturbation levels (8/255, 12/255, 16/255).
    Experiments are conducted on CIFAR10 \cite{krizhevsky2009learning} with PreactResNet18 \cite{he2016deep}. The total training duration is 110 epochs, with the learning rate decaying by a factor of 0.1 at the 100th and 105th epochs.}
    \label{tab:adversarial_performance}
    % \resizebox{\textwidth}{!}{
    \begin{tabular}{l|ccc|ccc|ccc}
    \toprule[1pt]
    
    \multirow{2}{*}{\textbf{Method}} & \multicolumn{3}{c|}{\(\epsilon = 8/255\)}& \multicolumn{3}{c|}{\(\epsilon = 12/255\)}& \multicolumn{3}{c}{\(\epsilon = 16/255\)}\\
    \cline{2-10}
    &\textbf{Clean Acc} & \textbf{PGD50} & \textbf{AutoAttack} & \textbf{Clean Acc} & \textbf{PGD50} & \textbf{AutoAttack} & \textbf{Clean Acc} & \textbf{PGD50} & \textbf{AutoAttack} \\
    \hline
    PGD-10 \cite{rice2020overfitting} & $80.55 \pm 0.37$ & $50.67 \pm 0.40$ & $46.37 \pm 0.76$ & $72.37 \pm 0.31$ & $38.60 \pm 0.39$ & $33.13 \pm 0.28$ & $67.20 \pm 0.69$ & $29.34 \pm 0.18$ & $21.98 \pm 0.30$ \\
    \hline
    GAT \cite{sriramanan2020guided}&- &- &- & $78.52 \pm 0.25$ &- & $0.54 \pm 0.53$ &- &- & \\
    \hline
    GAT+ELLE \cite{abad2024efficient} &- &- &- & $65.71 \pm 2.48$ &- & $13.83 \pm 4.63$ &- &- & \\
    \hline
    DOM-AT\cite{lin2023over}&$83.49 \pm 0.69 $&-& $48.41 \pm 0.28$& -& -& -& -& -& -\\
    \hline
    FreeAT \cite{shafahi2019adversarial}
    & $76.20 \pm 1.09$ & $43.74 \pm 0.41$ & $40.13 \pm 0.39$ & $68.07 \pm 0.38$ & $33.14 \pm 0.62$ & $27.65 \pm 0.38$ & $45.84 \pm 19.07$ & $0.00 \pm 0.00$ & $0.00 \pm 0.00$ \\
    \hline
    Zerograd \cite{golgooni2021zerograd}& $81.60 \pm 0.16$ & $47.56 \pm 0.16$ & $44.76 \pm 0.02$ & $77.52 \pm 0.21$ & $27.34 \pm 0.09$ & $29.88 \pm 0.23$ & $79.65 \pm 0.17$ & $6.37 \pm 0.23$ & $19.07 \pm 0.28$ \\
    \hline
    Multigrad \cite{golgooni2021zerograd}& $81.65 \pm 0.16$ & $47.93 \pm 0.18$ & $44.19 \pm 0.10$ & $81.09 \pm 4.67$ & $9.95 \pm 16.97$ & $0.00 \pm 0.00$ & $82.98 \pm 3.30$ & $0.00 \pm 0.00$ & $0.00 \pm 0.00$ \\
    \hline
    Gradalign \cite{andriushchenko2020understanding} & $82.10 \pm 0.78$ & $47.77 \pm 0.58$ & $44.66 \pm 0.21$ & $74.17 \pm 0.55$ & $34.87 \pm 1.00$ & $28.63 \pm 0.34$ & $60.37 \pm 0.95$ & $27.90 \pm 1.01$ & $17.46 \pm 1.71$ \\
    \hline
    % N-FGSM-ELLE &- & $45.05 \pm 0.26$ &- &- & $30.61 \pm 0.34$ & $61.21 \pm 0.14$ &- & $20.48 \pm 0.57$ &-\\
    FGSM-RS \cite{wong2020fast} & $83.91 \pm 0.21$ & $46.01 \pm 0.18$ & $42.88 \pm 0.09$ & $66.46 \pm 22.8$ & $0.00 \pm 0.00$ & $0.00 \pm 0.00$ & $66.54 \pm 12.25$ & $0.00 \pm 0.00$ & $0.00 \pm 0.00$ \\
    \hline
    N-FGSM \cite{de2022make}& $80.48 \pm 0.21$ & $47.91 \pm 0.29$ & $44.81 \pm 0.18$ & $71.30 \pm 0.12$ & $36.23 \pm 0.10$ & $30.56 \pm 0.12$ & $62.96 \pm 0.74$ & $27.14 \pm 1.44$ & $20.59 \pm 0.47$ \\
    \hline
    FGSM-MEP \cite{jia2022prior}&$83.57\pm 0.06$&$49.78\pm 0.15$&$45.28\pm 0.15$&$74.43 \pm 4.59$&$34.90\pm 2.08$&$27.50\pm 1.83$&$82.47\pm 5.85$&$0.47\pm 0.26$&$0.00\pm 0.00$\\
    \hline
    DOM-FGSM \cite{lin2023over} &$84.52 \pm 0.28$&- &$42.53 \pm 1.41$& -& -& -& -& -& -\\
    \hline
    RS-LAP \cite{lin2024layer}& $84.12 \pm 0.29$ &- & $43.14 \pm 0.45$ & $74.10 \pm 0.31$ &- & $26.04 \pm 1.04$ & $64.83 \pm 0.29$ &- & $15.69 \pm 0.28$ \\
    \hline
    N-LAP \cite{lin2024layer}& $80.76 \pm 0.15$ & $44.97 \pm 0.24$ &- & $71.91 \pm 0.19$ & $30.60 \pm 0.27$ &- & $63.73 \pm 0.27$ & $19.55 \pm 0.18$ & -\\
    \hline
    RS-AAER \cite{lin2024eliminating}& $83.83 \pm 0.27$ & $46.14 \pm 0.02$ &- & $74.40 \pm 0.79$ & $32.17 \pm 0.16$ &- & $64.56 \pm 1.45$ & $23.87 \pm 0.36$ & \\
    \hline
    N-AAER \cite{lin2024eliminating}& $80.56 \pm 0.35$ & $48.31 \pm 0.23$ &- & $71.15 \pm 0.18$ & $36.52 \pm 0.10$ &- & $61.84 \pm 0.43$ & $28.20 \pm 0.71$ & \\
    \hline
    FGSM-PBD \cite{zhao2024catastrophic}&$83.38\pm 0.19$ &$50.19\pm 0.09$&$45.45 \pm 0.12$ &$77.39 \pm 0.28$& $37.21\pm 0.33$& $30.37\pm 0.38$& $70.69\pm 0.21$&$27.39\pm 0.15$&$17.90\pm 0.24$\\
    \bottomrule[1pt]
    \end{tabular}
    % }
    % \vspace{-2mm}
\end{table*}

% encompasses various techniques such as adversarial fine-tuning and adversarial robust distillation, 
$\circ$ \textit{Loss functions in finetuned AT.} 
Li et al. \cite{li2017learning} propose a lifelong learning strategy that fine-tunes pre-trained models with training data of the target task while maintaining the model performance on the source task. This strategy minimizes the following objective function:
\[
\min_{{\theta}_s, {\theta}_t} \left[ \ell_{\text{source}}(\hat{y}_s, y_s ;\theta_s) + \ell_{\text{ce}}(x_t ,y_t;\theta_t) \right],
\]
where \(\theta_s\) and \(\theta_t\) denote the task-specific parameters for the source and target tasks, respectively. \(\ell_{\text{source}}\) is a modified cross-entropy loss emphasizing small prediction probabilities:
\[
\ell_{\text{source}}(\hat{y}_s, y_s;\theta_s) = -\sum_{i=1}^{C} y_{s,i}^\prime \log \hat{y}_{s,i}^\prime,
\]
where \(C\) is the number of categories. \(y_{s,i}^\prime\) and \(\hat{y}_{s,i}^\prime\) are derived from the ground-truth probability \(y_{s,i}\) and the predicted probability \(\hat{y}_{s,i}\), respectively,
\[
y_{s,i}^\prime = \frac{(y_{s,i})^{1/T}}{\sum_j (y_{s,j})^{1/T}}, \quad \hat{y}_{s,i}^\prime = \frac{(\hat{y}_{s,i})^{1/T}}{\sum_j (\hat{y}_{s,j})^{1/T}}.
\]
Here, the default value of \(T\) is set to 2.
% \[
% \langle \mathbf{P}, \mathbf{D}_{st} \rangle - \varepsilon \cdot h(\mathbf{P})) + \tau_1 \cdot \text{KL}(\mathbf{P}\mathbf{1} \| w_p) + \tau_2 \cdot \text{KL}(\mathbf{P}^\top\mathbf{1} \| w_t),
% \]
% where \(\langle\cdot\rangle\) represents the inner product, \(\mathbf{D}_{st}\) denotes the distance between samples in the source and target datasets, \(\tau_1\) and \(\tau_2\) are hyperparameters, and \(h(\cdot)\) is the entropy function. \(w_p\) and \(w_t\) are the category weights for the pre-training and target datasets, respectively. A higher value of \(\mathbf{P}\mathbf{1}\) indicates greater similarity of category in the pre-training dataset to the target dataset.
Zhu et al. \cite{zhu2023improving} identify and refine non-robust key modules, where the robust criticality of module \(i\) is calculated by
\[
\max_{\theta' \in \mathcal{C}_\theta} \frac{\sum_{(x, y) \in D} \max_{\delta} \ell_\text{ce}(x + \delta, y; \theta+\theta^\prime)}{\sum_{(x, y) \in D} \max_{\delta} \ell_\text{ce}(x + \delta, y; \theta)},
\]
% where \( \mathcal{S} \) denotes the set of allowable perturbations. 
where \( \theta' = \{0, \ldots, 0, \theta_i^\prime, 0, \ldots, 0\} \) represents the weight perturbation applied to the module \(i\), and
\[
\mathcal{C}_\theta = \left\{ \theta' \;\middle|\; \|\theta'\|_p \leq \varepsilon\cdot \|\theta_{i}\|_p \right\}.
\]

$\circ$ \textit{Loss functions in robust distillation.} 
Conventional knowledge distillation methods \cite{gou2021knowledge,cho2019efficacy,park2019relational} are commonly expressed as follows:
\[
\min_\theta [\lambda \cdot \ell_\text{ce}(S(x;\theta), y) + (1-\lambda) \cdot \text{KL}(S(x;\theta)\|T(x))],
\]
where \( \lambda \in [0, 1] \) controls the regularization strength. 
\(T(\cdot)\) and \(S(\cdot;\theta)\) are teacher and student models, respectively.

In addition to inheriting the performance of teacher models on the original task, robust distillation \cite{shafahi2019adversarially} requires the transfer of robust features to student models.
Goldblum et al. \cite{goldblum2020adversarially} implement robust distillation by minimizing
\begin{equation}\label{Goldblum}
    \lambda \cdot \ell_\text{ce}\left( S_\tau(x;\theta), y \right) + (1 - \lambda) \cdot\tau^2 \cdot \mathrm{KL}\left( S_\tau(x + \delta;\theta)\| T_{\tau}(x) \right),
\end{equation}
where \(\tau\) is the temperature of Softmax. The adversarial perturbations \(\delta\) are calculated by
\[ \delta = \arg \max_{\mathcal{B}(x, \epsilon)} \ell_\text{ce}\left( S_\tau(x + \delta;\theta), y \right) .\]
Zi et al. \cite{zi2021revisiting} produce adversarial perturbations based on soft labels generated by the teacher model,
\[
\delta = \arg\max_{\mathcal{B}(x, \epsilon)} \mathrm{KL}(S(x+\delta; \theta) \| T(x)).
\]
Maroto et al. \cite{maroto2022benefits}) propose the label mixing strategy, replacing \(\ell_\text{ce}(S_\tau(x; \theta), y)\) in Eq. (\ref{Goldblum}) with 
\[
\ell_\text{ce}(S(x+\delta;\theta), \beta\cdot T(x+\delta) + (1-\beta)\cdot y),
\]
where \(\beta \in [0, 1]\) controls the mixture degree of distilled and ground-truth labels.
Huang et al. \cite{huang2023boosting} replace \(\ell_\text{ce}(S_\tau(x;\theta), y)\) in Eq. (\ref{Goldblum}) with KL divergence:
\[
\lambda \cdot\mathrm{KL}(S(x;\theta) \| T(x)) + (1-\lambda)\cdot\mathrm{KL}(S_\tau(x+\delta;\theta) \| T_\tau(x+\delta)).
\]
% Additionally, based on the information bottleneck theory,  
Kuang et al. \cite{kuang2024improving} introduce an information bottleneck distillation method, replacing one-hot labels with robust prior information from pre-trained models,
\[
\lambda\cdot\ell_{ce}(S(x;\theta), T(x)) + (1-\lambda)\cdot \ell_{ce}(S(x+\delta;\theta), T(x)). \]

$\circ$ \textit{Loss functions in curriculum AT.}
To improve worst-case robustness on large datasets, Cai et al. \cite{cai2018curriculum} propose curriculum AT, which progressively increases attack intensity during training. Specifically, after each training epoch, the \(k\)-accuracy is evaluated on a selected validation set \(V\), 
\[
\frac{| \{ (x, y) \in V \mid \forall i \in \{0, \dots, k\}, f(\text{PGD}_i(x);\theta) = y \} |^\dagger}{|V|^\dagger}.
\]
Here, \(|\cdot|^\dagger\) calculates the number of samples, and \(\text{PGD}_i(\cdot)\) represents a PGD attack with \(i\) steps. 
The attack strength is increased if the \(k\)-accuracy does not improve over the past 10 epochs. 
Kinfu et al. \cite{kinfu2022analysis} propose an adaptive method for distributing perturbation budgets based on the calculated probability across different attacks,
\[
P(\epsilon_i) = \frac{\sum_{n=1}^{N} \ell_{ce}(x_n + \delta^i_n, y_n; \theta)}{\sum_{j=1}^{K} \sum_{n=1}^{N} \ell_{ce}(x_n + \delta^j_n, y_n;\theta)},
\]
where \(\delta^i_n\) is the perturbation applied to \(x_n\) constrained by a budget \(\epsilon_i\). \(K\) denotes the number of discrete attack budgets.

$\circ$ \textit{Loss functions in domain AT.}
Domain AT \cite{ganin2016domain} involves two domains, source and target, aiming to improve generalization performance across domains.

To optimize the model weights, Liu et al. \cite{liu2019transferable} simultaneously minimize the domain discrimination loss \(\ell_d\) and the source domain classification loss \(\ell_c\):
\[
\ell_{d} = -\frac{1}{n_s}\sum_{i=1}^{n_s}\log\left[\text{D}(x_{s,i})\right] -\frac{1}{n_t}\sum_{i=1}^{n_t}\log\left[1 - \text{D}(x_{t,i})\right],
\]
\[
\ell_{c} = \frac{1}{n_s}\sum_{i=1}^{n_s}\ell(x_{s,i}, y;\theta) + \frac{1}{n_t}\sum_{i=1}^{n_t}\|f(x_{t,i};\theta) - f(x_{i};\theta)\|_p,
\]
where \(n_s\) and \(n_t\) denote the number of examples in the source and target domains, respectively. \(\text{D}(\cdot)\) represents the domain discriminator. \(x_s\) and \(x_t\) are generated adversarial examples:
\[
x_s = x + \delta_s, \quad x_t = x + \delta_t,
\]
\[
\delta_s = \delta_s + \alpha\cdot (\nabla_{\delta_s}\ell_d + \nabla_{\delta_s}\ell_c) - \gamma\cdot \|\delta_s\|_2,
\]
\[
\delta_t = \delta_t + \alpha \cdot\nabla_{\delta_t}\ell_d  - \gamma\cdot \|\delta_t\|_2,
\]
where \(\alpha\) and \(\gamma\) are hyper-parameters.
Smooth domain AT  \cite{rangwani2022closer} alternately updates the discriminator and the adversarially trained model. Specifically, the discriminator weights are updated by maximizing the domain discrimination loss, while the adversarial model weights are optimized by minimizing the task and domain discrimination losses.

$\circ$ \textit{Balance Multiple Loss Terms.} Researchers also explore balancing various loss terms \cite{yu2022robust}.
For instance, Wallace et al. \cite{wallace2021analyzing} introduce dynamic hyperparameter learning based on the classification accuracy of clean samples. Tong et al. \cite{tong2024taxonomy} propose an epoch-by-epoch label relaxation method that adjusts regularization hyperparameters according to the classification confidence of adversarial examples.

\underline{\textit{2) Label Modification:}}
Wang et al. \cite{wang2019bilateral} assign non-zero random values to the non-target classes in one-hot labels. 
Chen et al. \cite{chen2020smoothing} employ soft labels generated by a teacher model for both clean and adversarial examples to distill the student model. 
Li et al. \cite{li2024soften} propose a self-guided label refinement method that generates interpolated soft labels for student models without relying on external teacher models.  
Dong et al. \cite{dong2021exploring} attribute robust overfitting in AT to model memorization of one-hot labels, and address this by using momentum-based prediction statistics as labels for adversarial examples. 
Song et al. \cite{song2024regional} construct an adversarial region based on the initial and final adversarial examples, selecting adversarial points from this region and assigning soft labels based on their distance from the original examples.

\underline{\textit{3) Prediction Modification:}}
Zhou et al. \cite{zhou2022modeling} model the relationship between adversarial and ground-truth labels using an instance-dependent transition matrix, and convert adversarial predictions during inference. Stutz et al. \cite{stutz2020confidence} propose confidence-calibrated AT, where low-confidence samples are excluded from evaluation. Namely, the model is trained to reduce its prediction confidence as perturbation levels increase.

\underline{\textit{4) Weight-related Settings:}}

$\circ$ \textit{Adversarial weight perturbation.} 
Wu et al. \cite{wu2020adversarial} observe a positive correlation between robust generalization performance and model weight flatness. To enhance this flatness, they introduce adversarial perturbations to model weights. Similar approaches \cite{xing2021algorithmic, sankaranarayanan2018regularizing} propose applying adversarial perturbations to both inputs and model weights. 
To reduce the influence of weight perturbations on model predictions,
Park et al. \cite{park2021reliably} introduce perturbations only to specific sensitive feature layers. 
Lin et al. \cite{lin2024layer} present adaptive weight perturbations across layers to hinder the generation of pseudo-robust shortcuts. 
To mitigate robust overfitting, Yu et al. \cite{yu2022robust} introduce weight perturbations when the classification loss drops below a pre-defined threshold, while dynamically adjusting the hyperparameters of different loss components.

$\circ$ \textit{Random weight perturbation.} 
DeepAugment \cite{hendrycks2021many} applies random noise to the weights of image-to-image models (\textit{e.g.}, super-resolution). Similarly, Jin et al. \cite{jin2023randomized} introduce random noise into deterministic weights and apply a Taylor expansion to analyze losses calculated based on perturbed weights.

$\circ$ \textit{Weight standardization.}  
To facilitate robust feature learning during training, Pang et al. \cite{pang2020boosting} present a hypersphere embedding network that incorporates various normalization operations, such as feature normalization and weight normalization. 
Brock et al. \cite{brock2021characterizing} introduce a signal propagation analysis toolset for ResNet, replacing activation normalization layers with an adapted weight standardization technique.

$\circ$ \textit{Model weight selection.}  
Subspace AT \cite{li2022subspace} leverages Singular Value Decomposition to determine the most significant gradients during training.
Croce et al. \cite{croce2023seasoning} construct adversarially robust model soups, which are linear combinations of parameters for multiple models, effectively balancing robustness against different \(\ell_p\)-norm bounded adversaries.

\underline{\textit{5) Softmax Variants:}}
To address class-wise imbalances in adversarial robustness, Tian et al. \cite{tian2021analysis} introduce temperature-PGD in the inner maximization, which modifies back-propagated gradients by adjusting the Softmax temperature.
Hou et al. \cite{hou2023improving} argue that untargeted attacks often adopt the nearest target as the adversarial direction. 
To mitigate this, they adjust the Softmax temperature based on the cosine similarity between diverse categories. 
To enhance knowledge transfer efficiency, Wang et al. \cite{wang2024out} present an interactive temperature adjustment strategy, starting with a high temperature and gradually reducing it. They also introduce an adaptive generator that balances performance for clean and adversarial examples based on the model confidence in adversarial examples.

\underline{\textit{6) Optimization Algorithms:}}
Various optimization algorithms can be employed to update the model weights in AT.

$\circ$ \textit{Common optimizers.}  
Adagrad \cite{duchi2011adaptive} updates model weights by accumulating past gradients:
\[
\theta_{t+1} = \theta_t - \frac{\eta}{\sqrt{G_{t} + c}} \cdot g_t,
\]
where \(\theta_t\) and \(g_t\) represent the model parameters and gradients at the time step \(t\), respectively.  
\(\eta\) denotes the global learning rate, while \(G_{t}\) is the sum of previous squared gradients, with \(c\) 
being a small constant for numerical stability.
% being a small constant to prevent division by zero. 
RMSprop \cite{tieleman2017divide} enhances Adagrad by replacing \(G_{t}\) with an exponentially decaying average of squared gradients, \(E[g^2]_t\):
\[
\theta_{t+1} = \theta_t - \frac{\eta}{\sqrt{E[g^2]_t + c}} \cdot g_t.
\]
Adam \cite{kingma2014adam} combines the strengths of Adagrad and RMSprop by maintaining exponentially decaying averages of both past gradients (momentum) and squared gradients (adaptive learning rates), formulated as follows:
\[
\theta_{t+1} = \theta_t - \frac{\eta \cdot \hat{m}_t}{\sqrt{\hat{v}_t} + c},
\]
where \(\hat{m}_t\) and \(\hat{v}_t\) are the bias-corrected estimates of the first and second gradient moments, respectively.

$\circ$ \textit{Optimization algorithms in AT.} 
Zhang et al. \cite{zhang2022revisiting} utilize a bi-level optimization technique for fast AT methods, which does not rely on gradient-based approaches or explicit regularization constraints.
Wang et al. \cite{wang2022self} accelerate the optimization process by incorporating the momentum aggregation strategy.
Acuna et al. \cite{acuna2022domain} demonstrate that employing a gradient descent-based optimizer often limits the transfer performance in domain AT. They suggest using high-order ordinary differential equation solvers (\textit{e.g.}, Runge-Kutta) and provide asymptotic convergence guarantees.
Jin et al. \cite{jin2022enhancing} treat model weights as random variables and optimize them with a second-order statistic technique.

\underline{\textit{7) Learning Rate Schedules:}}
The learning rate schedule can impact the stability and effectiveness of AT techniques \cite{jia2022prior}.

Beyond employing a fixed learning rate, Bengio et al. \cite{bengio2012practical} propose an epoch-decay schedule, in which the learning rate decreases as training progresses:
\[
\eta_{e} = \frac{\eta_0}{1 + \lambda\cdot e},
\]
where \(\eta_e\) is the learning rate at epoch \(e\), and \(\lambda\) is the decay rate. Similarly, He et al. \cite{he2016deep} introduce a step-decay schedule,
\[
\eta_s = \eta_0 \cdot \gamma^{\left\lfloor \frac{s}{{I}} \right\rfloor},
\]
where \(\eta_s\) represents the learning rate at training step \(s\), \(\gamma\) is the decay factor, and \({I}\) denotes the interval for decaying the learning rate. 
Szegedy et al. \cite{szegedy2017inception} design an exponential learning rate schedule, formulated as
\[
\eta_t = \eta_0 \cdot \text{exp}({-\lambda t}).
\]
Loshchilov et al. \cite{loshchilov2016sgdr} present a cosine annealing approach, 
\[
\eta_t = \eta_{\text{min}} + \frac{1}{2} (\eta_{\text{max}} - \eta_{\text{min}}) \left(1 + \cos\left(\frac{t}{T_{\text{max}}} \pi\right)\right),
\]
where \(\eta_{\text{min}}\) and \(\eta_{\text{max}}\) denote the minimum and maximum learning rates, respectively, and \(T_{\text{max}}\) is the maximum iteration number. Smith et al. \cite{smith2017cyclical} propose a cyclical learning rate schedule, defined as:
\[
\eta_t = \eta_{\text{min}} + \frac{1}{2} (\eta_{\text{max}} - \eta_{\text{min}}) \left(1 + \cos\left(\frac{t \mod T}{T} \pi\right)\right).
\]
Here, \(T\) denotes the cycle period, and `mod' calculates the remainder of a division.

\section{Challenges and potential research directions}\label{sec4}

\subsection{Challenges}

\underline{\textit{\textit{1) Catastrophic Overfitting:}}}
Catastrophic overfitting refers to the phenomenon where models perform well on training data but experience a sharp decline in accuracy on unknown adversarial examples. 
Kang et al. \cite{kang2021understanding} observe that training with a single type of adversarial perturbation often leads to catastrophic overfitting.
To mitigate this, they propose leveraging diverse perturbations that span the full range of floating-point values. 
Golgooni et al. \cite{golgooni2021zerograd} argue that small backpropagated gradient values can trigger catastrophic overfitting and recommend removing small gradient elements. 
Tsiligkaridis et al. \cite{tsiligkaridis2022understanding} observe a close relationship between catastrophic overfitting and small distortions, introducing a dynamic attack intensity adjustment method based on prior distortion levels. 
Huang et al. \cite{huang2023fast} emphasize that fitting instances with large gradient norms is prone to catastrophic overfitting and thus present adaptive AT, where the attack stride is inversely proportional to the instance gradient norm. 
He et al. \cite{he2023investigating} explore catastrophic overfitting from a self-fitting perspective, \textit{i.e.}, the model overfits to specific adversarial perturbations used during training and loses its generalization ability to slightly different perturbations. 
They further find that catastrophic overfitting results in significant channel separation.

\underline{\textit{\textit{2) Fairness:}}}
Xu et al. \cite{xu2021robust} observe that robustness improvements vary across different model layers and propose adjusting model weights based on the layer-level robustness. 
To improve adversarial robustness for the worst-performing class while preserving robustness across other classes, Li et al. \cite{li2023wat} propose a worst-class AT framework. This framework includes rigorous theoretical analyses, such as a generalization error bound for worst-class robust risk, and introduces a metric to assess both average and worst-class accuracies.
Sun et al. \cite{sun2023improving} identify two fairness issues in AT: the varying difficulty in generating adversarial examples for different classes (source-class fairness) and the unbalanced targets of generated adversarial examples (target-class fairness). For source-class fairness, they generate adversarial examples near the decision boundary, such as the first adversarial example from PGD. To address target-class fairness, they introduce uniform distribution constraints for both the final clean sample and the first adversarial example. To improve fairness in transferable AT, Liu et al. \cite{liu2022improved} propose an unbalanced optimal transport method, which utilizes data categories from the source dataset that closely match the distribution of the target dataset.

\underline{\textit{3) Performance Trade-off:}}
Recent studies reveal that enhancing adversarial robustness often compromises the prediction performance for clean samples. To address this trade-off, Wang et al. \cite{wang2020once} introduce hyperparameters into BN layers, enabling the adjustment of prediction focus during inference. 
Mehrabi et al. \cite{mehrabi2021fundamental} propose a Pareto optimal trade-off curve for explaining AT performance, analyzing how factors such as feature correlation, attack intensity, and network width affect this curve. Ge et al. \cite{ge2023advancing} examine the role of data types and propose unifying various AT strategies. They demonstrate that challenges like performance trade-offs, robust overfitting, and catastrophic overfitting, can be alleviated by incorporating different types of training examples. 

\underline{\textit{4) Time-Efficiency:}}
Although AT techniques effectively defend against adversarial attacks, they bring considerable computational overhead. 
To mitigate this, Shafahi et al. \cite{shafahi2019adversarial} propose returning both weight and input gradients within each backpropagation step. 
Zhang et al. \cite{zhang2019you} recalculate gradients only for the first layer, terminating forward and backward propagation for the remaining layers.
Zheng et al. \cite{zheng2020efficient} reuse adversarial examples generated in earlier epochs, as they remain effective in later training stages.
Rocamora et al. \cite{rocamora2024efficient} propose a local linearity regularization to avoid the computational overhead caused by the double backpropagation of regularization terms. 
More recently, researchers have focused on developing fast AT methods \cite{jia2024fast}.

% \[ v_t = \beta v_{t-1} + (1 - \beta)\|\nabla_x\ell(x+\delta_{t-1};\theta)\|_2^2\]
% \[\alpha_{t} = \gamma/(c+\sqrt{v_t})\]
% \[x_t = \text{Proj}_{B(x,\epsilon)}(x_{t-1} + \alpha_t\text{sign}(\nabla_x\ell(x+\delta_{t-1},y;\theta)))\]
%Notably, AT may encounter catastrophic overfitting \cite{zhao2024catastrophic,lin2024eliminating,ge2023advancing,jiang2023towards}, where the model's robustness to adversarial attacks drastically diminishes to nearly zero during training. (1) The overfitting partially occurs because adversarial perturbations are more homogeneous than original examples, causing the model to quickly overfit to these perturbations \cite{araujo2019robust,wong2020fast,wang2019bilateral}.  This can be mitigated by utilizing aforementioned data augmentation techniques. (2) Furthermore, under the same setting, the severity of adversarial attacks can vary significantly across different examples \cite{tao2021better,li2023wat}, or even across different locations within the same example \cite{yu2022robust,sankaranarayanan2018regularizing}. Consequently, models may overfit to data with lower levels of adversarial attack severity, losing robustness to more severe adversarial perturbations. Existing work resolves this by calculating neighboring loss differences (or backpropagated gradient values) to filter out data with low attack severity \cite{zhao2023fast} (or low gradient values \cite{li2022subspace,huang2023fast,golgooni2021zerograd}), thereby ensuring model robustness across a broader spectrum of perturbations.

\subsection{Potential Research Directions}

\underline{\textit{1) Unexplored Fields:}}
In the supplementary material, we summary the application of AT across 100 popular deep learning tasks. Among them, 18 areas remain unexplored regarding the feasibility of integrating AT, including image deblurring, object tracking, face shape modeling, optical flow, virtual reality, environmental sound classification, motion deblurring, audio retrieval, 3D instance segmentation, 3D super-resolution, 3D object tracking, 3D style transfer, speaking style conversion, audio denoising, audio restoration, scene graph generation, and personalized medicine.
Researchers can investigate whether AT can enhance these fields by analyzing its potential benefits and challenges.

\underline{\textit{2) What Benefits can Catastrophic Overfitting Offer to AT?}}
As shown by Zhao et al. \cite{zhao2024catastrophic}, catastrophic overfitting in AT can address the performance trade-off between clean and adversarial examples. However, simply introducing random noise to inputs of overfitted models fails to defend against adaptive attacks. Therefore, researchers can develop techniques that address both the performance trade-off and the vulnerability of networks to adaptive attacks.

Moreover, catastrophic overfitting could be leveraged for data copyright protection, \textit{e.g.}, ensuring that protected datasets are used solely for designated tasks. Researchers could design an adversarial dataset that induces catastrophic overfitting when applied to non-specified tasks.

\underline{\textit{3) Unified Fairness Framework:}}
Recent AT research often focuses on individual fairness issues, such as inter-layer, inter-class, or inter-sample robustness fairness. Moreover, adversarially trained models face fairness challenges in their predictions for the original task. Hence, a unified fairness framework can be developed to address these concerns hierarchically, from coarse-grained to fine-grained levels.

\underline{\textit{4) Interpretability and Model Design:}}
Previous work typically minimizes the distributional distance between clean and adversarial examples and shows a performance trade-off between these two types of examples. 

Assuming that each network can learn only a single distribution, the trade-off issue becomes plausible, as adversarially trained models are required to learn from both clean and adversarial distributions. Researchers can either validate this assumption or develop models capable of learning from both distributions. These models should automatically handle three tasks: adversarial example detection, adversarial example classification, and clean sample classification.

For instance, Xie et al. \cite{xie2019intriguing} employ separate BN layers for clean and adversarial examples. However, this approach is primarily analytical and challenging to implement during inference, as it requires specifying the sample type (clean or adversarial) to select the correct BN layer. To overcome this, an adaptive BN plug-in can be designed to automatically select the appropriate BN layer based on the input type.

Researchers can also construct an adversarial dataset that closely matches the original data distribution. For example,  distribution similarity metrics like Wasserstein distance \cite{panaretos2019statistical} and maximum mean discrepancy \cite{smola2006maximum}, can be applied to generate adversarial examples. 
% These samples should effectively deceive trained models while maintaining a close resemblance to the original data distribution.

\underline{\textit{5) Resource-consumption Consideration:}} 
Compared to standard model training, AT techniques demand higher computational resources, especially when training large models on massive datasets. To address this, integrating fast AT with network pruning provides a potential solution. 
By constructing several pruned models from the trained large model and computing adversarial examples for these pruned models in parallel, researchers can identify the most aggressive adversarial example, which can then be applied to train the large model.

\section{Conclusion}
This paper presents a comprehensive review of recent advancements in AT.
We begin by illustrating the implementation of AT techniques and their practical applications in real-world scenarios.
Next, we construct a unified algorithmic framework for AT and categorize AT methods across three dimensions: data enhancement, network architecture, and training configurations. Additionally, we summarize common challenges in AT and identify several potential directions for future research.
%\section*{Acknowledgements}
%This work was supported by the National Natural Science Foundation of China under Grant 62276046 and by Dalian Science and Technology Innovation Foundation under Grant 2023JJ12GX015.

\bibliographystyle{IEEEtran}
\bibliography{main}

% \begin{IEEEbiographynophoto}{Jane Doe}
% Biography text here without a photo.
%\end{IEEEbiographynophoto}

%\begin{IEEEbiography}[{\includegraphics[width=1in,height=1.25in,clip,keepaspectratio]{fig1.png}}]{IEEE Publications Technology Team}
%In this paragraph you can place your educational, professional background and research and other interests.\end{IEEEbiography}

\end{document}